\documentclass[11pt, a4paper, logo, copyright]{googledeepmind}

\usepackage{gdm-colors}
\usepackage{mathtools}
\usepackage{bm}
\usepackage{multirow}
\usepackage{subcaption}
\usepackage{placeins}
\usepackage[nameinlink]{cleveref}
\usepackage{csquotes}
\usepackage{tikz}
\usetikzlibrary{arrows.meta}

\usepackage[
  style=authoryear-comp,
  sorting=ynt,
  natbib=true,
  backend=biber,
  url=false,
  doi=false,
  maxcitenames=2,
  maxbibnames=99,
  uniquelist=false,
  uniquename=false,
  giveninits=true,
  dashed=false
]{biblatex}
\let\cite\textcite

\DeclareFieldFormat{citehyperref}{%
  \DeclareFieldAlias{bibhyperref}{noformat}%
  \bibhyperref{#1}}
\DeclareFieldFormat{textcitehyperref}{%
  \DeclareFieldAlias{bibhyperref}{noformat}%
  \bibhyperref{%
    #1%
    \ifbool{cbx:parens}
      {\bibcloseparen\global\boolfalse{cbx:parens}}
      {}}}
\savebibmacro{cite}
\savebibmacro{textcite}
\renewbibmacro*{cite}{%
  \printtext[citehyperref]{%
    \restorebibmacro{cite}%
    \usebibmacro{cite}}}
\renewbibmacro*{textcite}{%
  \ifboolexpr{
    ( not test {\iffieldundef{prenote}} and
      test {\ifnumequal{\value{citecount}}{1}} )
    or
    ( not test {\iffieldundef{postnote}} and
      test {\ifnumequal{\value{citecount}}{\value{citetotal}}} )
  }
    {\DeclareFieldAlias{textcitehyperref}{noformat}}
    {}%
  \printtext[textcitehyperref]{%
    \restorebibmacro{textcite}%
    \usebibmacro{textcite}}}

\newcommand{\fourbrevision}[1]{{#1}}

\graphicspath{{figures/}}

\title{Kwai Summary Attention V2 Technical Report}
\author{\large OneRec Team}
\renewcommand{\today}{}
\hypersetup{
  pdftitle={Kwai Summary Attention V2 Technical Report},
  pdfsubject={Working draft: Summary KV compression and Summary-guided sparse attention},
  pdfkeywords={KSA, C-KSA, S-KSA, long-context language models, KV cache, block sparse attention}
}

\begin{abstract}
Summary-token architectures such as Kwai Summary Attention (KSA) compress distant context into learnable Summary token per chunk, and hybrid variants interleave these layers with full attention to retain exact access to the original text.
In such hybrid KSA-Full architecture, the interleaving inserted summary tokens are served as a streaming flow role across different layers.
Although effective, however, it also introduce some bottlenecks: (1) the different layers summary
compression chunk must be same, therefore hard to reach a flexible KV Cache compression, e.g., different layer set different summary compression chunk window.
(2) in the full-attention layers, the summary token introduced additional computation cost, but such layer ignore them and still pay quadratic prefilling, therefore unfriendly to the computation side.
Motivated by them, we developed the KSA V2, which not only achieves a more flexible KV cache memory compression but also reduces the computation cost in Full attention layer.
Specifically, we found that the summary tokens can serve two roles: the KV memory and sparse indexer.
As KV memory, they can be further compressed without coarsening the representation passed between layers: \textbf{Compressed KSA (C-KSA)} pools groups of Summary KV into layer-local Meta entries, raising a larger chunk distant history compression.
As sparse index, the summary token could conduct exact-access layers where to look: \textbf{Sparse KSA (S-KSA)} builds block-level routing keys from the Summaries and attends only to selected original-text blocks.
We conduct experiments based on the 1.9B (MHA) and 4B (GQA) Hybrid KSA backbones, and we mostly adopt them by training the new compressed and sparse modules.
In 1.9B (MHA) and 128K sequence scale, it reduces resident cache size by 40.4\% and achieves $1.51\times$ prefill and $1.27\times$ decode throughput.
\end{abstract}

\begin{document}
\maketitle
\begin{figure}[!b]
 \centering
 \captionsetup{font=small}
 \includegraphics[width=\linewidth]{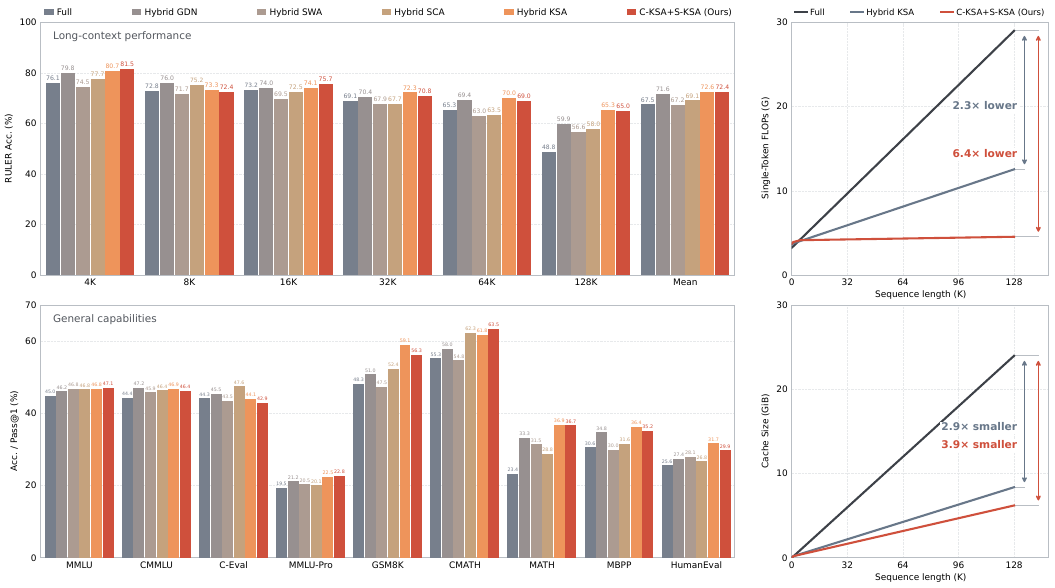}
 \caption{Left: long-context and general capabilities of \fourbrevision{1.9B} C-KSA+S-KSA and the reference models reported in KSA.
 Right: per token inference FLOPs and cache size ($1\mathrm{GiB}=2^{30}$ bytes).}
 \label{fig:quality-efficiency}
\end{figure}

\clearpage

\section{Introduction}
\label{sec:introduction}

Serving long contexts incurs substantial storage and computation costs.
With full attention, the key--value (KV) cache grows linearly with context length, prefill attention is quadratic, and each decoding query reads the full history.
Per-token KV compression~\citep{rw_gqa,rw_deepseekv2} reduces storage per position but retains an entry for every token.
Sparse attention~\citep{rw_moba,rw_nsa,rw_deepseekv32} reduces each query's read set while retaining token-level KV.
Sliding-window and recurrent linear attention~\citep{rw_mistral,rw_gated_deltanet} reduce both storage and computation, but do not preserve direct access to distant tokens.
Hybrid architectures balance these trade-offs by interleaving compact-memory layers with full-attention layers that retain direct access to the original text.

\begin{figure}[t]
  \centering
  \includegraphics[width=\linewidth]{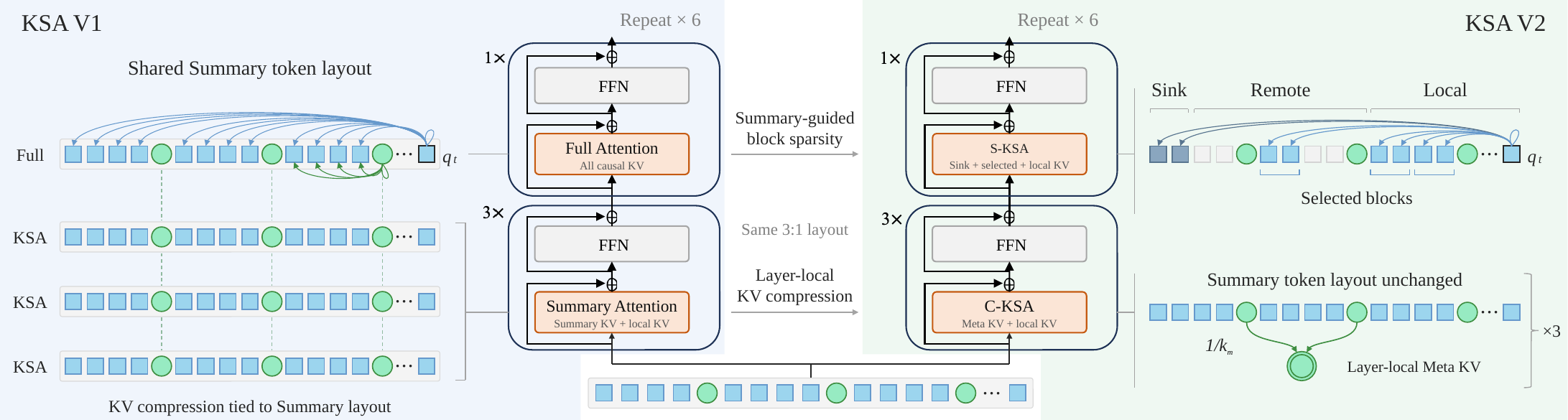}
  \caption{Architecture overview: from Hybrid KSA to KSA V2.}
  \label{fig:overview}
\end{figure}

Summary-token hybrids are a strong instance of this design.
Kwai Summary Attention (KSA)~\citep{rw_ksa} appends a learned Summary token after every $k_s$ text tokens; in KSA layers, text queries read nearby chunks at full resolution and distant chunks through their Summaries.
Hybrid KSA interleaves three such layers with one full-attention layer (Figure~\ref{fig:overview}) and retrieves long-range information better than a same-size full-attention model.

In these hybrids, however, the Summaries are underused.
They are computed and refined at every layer, yet serve only as low-resolution memory for KSA layers, leaving an inefficiency on each side.
On the efficient side, compression is tied to chunk size: a KSA layer stores one entry per distant chunk, so compressing further requires larger chunks, which also coarsen the Summaries that every later layer reads and refines.
\fourbrevision{The KSA report's 4B ablation shows that coarser Summary chunks can degrade retrieval performance.}
On the exact side, full-attention layers read every token, although most barely matter.
\fourbrevision{At 128K, oracle block selection restricts the main-attention read set to about 4\% of text positions, yet retains about 82\% of decode attention mass and near-dense retrieval performance in a frozen Hybrid KSA (Figure~\ref{fig:routing-budget}; Table~\ref{tab:oracle-quality}).}
What is missing is a cheap way to know where to look, and the Summaries, which describe every chunk, are never consulted.

We argue that a single summary hierarchy can play both roles.
As memory, Summaries can be compressed a second time when stored, without coarsening the representation that is propagated; as an index, they can direct exact-access layers to the few original-text blocks worth reading.
\textit{Efficient layers store coarse, exact layers read fine, and the Summaries connect the two.}

We instantiate this idea with two drop-in layer types, which together form \textbf{KSA V2}.
\textbf{Compressed KSA (C-KSA)} pools every $k_m$ consecutive Summary KV entries into one layer-local Meta entry, using learned dimension-wise weights as in DeepSeek-V4~\citep{rw_deepseekv4}.
Meta entries are never fed forward, so the Summary sequence keeps its granularity while distant-history storage shrinks by a factor of $k_s k_m$ ($1024\times$ with $k_s=8$ and $k_m=128$).
\textbf{Sparse KSA (S-KSA)} replaces full attention: each Summary pools the tokens of each $b$-token block in its chunk into a routing key, a Lightning Indexer~\citep{rw_deepseekv32} selects the top blocks, and the main attention reads their original KV with attention sinks and a local window.
Routing needs $b\times$ fewer candidates than token-level indexing in DSA, and unlike CSA~\citep{rw_deepseekv4}, the main attention never reads compressed text.

Because both layer types reuse existing Summaries, neither adds tokens nor changes the Summary layout, so both can be retrofitted into a pretrained hybrid.
\fourbrevision{The adapted 1.9B model largely preserves Hybrid KSA's retrieval and general capabilities while outperforming Full on retrieval.}
\fourbrevision{Ablations show that C-KSA alone matches its parent and that sparse adaptation incurs a smaller retrieval loss on the compressed backbone than on the original (Section~\ref{sec:model-quality}).}
\fourbrevision{At 128K, the model reduces resident cache size by 40.4\% and achieves $1.51\times$ prefill and $1.27\times$ decode throughput relative to Hybrid KSA. Compared with Full, it reduces cache size by 73.7\% and achieves $2.16\times$ prefill throughput (Section~\ref{sec:inference-efficiency}).}
The remaining cache is mostly original-text KV in the exact-access layers, and aggregation-heavy tasks such as frequent-word extraction degrade under block selection (Table~\ref{tab:niah_subtasks}).

\fourbrevision{The approach also retains near-parent retrieval performance on a 4B GQA backbone, supporting transfer beyond the 1.9B MHA setting.}

Our contributions are threefold:
\begin{itemize}
  \item We propose C-KSA, which compresses Summary KV at storage time into layer-local Meta entries, so that further compression of distant history no longer requires coarser Summaries.
  \item We propose S-KSA, which turns existing Summaries into block-level routing keys for sparse attention over selected original-text blocks.
  \item We provide a retrofit recipe and GPU kernels that convert a pretrained Hybrid KSA model with RULER performance comparable to its parent, and we report where the remaining cost lies and where the approach fails.
\end{itemize}

\section{Methodology}
\label{sec:methodology}
\label{sec:method-overview}

KSA V2 uses the Summaries in two roles: C-KSA compresses them further as long-range memory, and S-KSA uses them as an index to select original-text blocks for sparse attention.
\fourbrevision{As shown in Figure~\ref{fig:overview}, we retain Hybrid KSA's $3{:}1$ layout, repeating three C-KSA layers followed by one S-KSA layer throughout the backbone.}

We denote text tokens by $T_i$, chunk Summaries by $S_c$, and Meta KV entries by $M_g$.
The chunk size $k_s$ specifies the number of text tokens per Summary, $k_m$ the number of Summaries per Meta entry, and $b$ the number of text tokens per routing block.
For a text query at position $t$, $q_t$ and $O_t$ denote the attention query and output, respectively.


\subsection{Compressed KSA (C-KSA)}
\label{sec:method-cksa}

\begin{figure}[t]
  \centering
  \includegraphics[width=\linewidth]{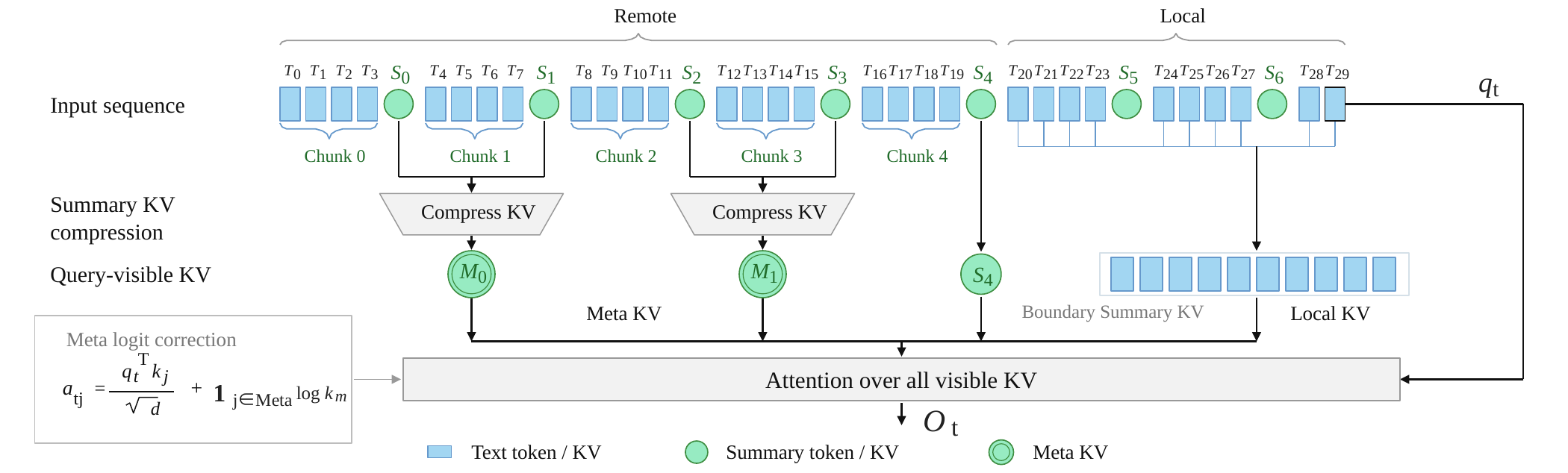}
  \caption{Compressed KSA with $k_s=4$ and $k_m=2$. Complete remote Summary groups are compressed into Meta KV, while boundary Summary KV and local text KV remain directly accessible.}
  \label{fig:cksa-hierarchy}
\end{figure}

C-KSA adds a second level of compression to KSA: KSA inserts one Summary token after every $k_s$ text tokens, and C-KSA aggregates every $k_m$ consecutive Summary KV entries into one Meta KV entry.
This increases the compression factor for distant history from $k_s$ to $k_s k_m$, while keeping recent text directly accessible (Figure~\ref{fig:cksa-hierarchy}).

\paragraph{Text-to-Summary aggregation.}
As in KSA, we partition the text sequence into chunks of $k_s$ tokens and append a Summary token $S_c$ to each chunk $c$.
All Summary tokens share a learned initial embedding and use the same attention and feed-forward parameters as text tokens.
At each layer, $S_c$ attends to its chunk's text tokens and itself to aggregate chunk-level information.
Text tokens retain their original positional indices, and $S_c$ uses the position of its chunk's last text token.

\paragraph{Summary-to-Meta aggregation.}
In each C-KSA layer, we divide the Summary sequence into non-overlapping groups of $k_m$ consecutive Summaries.
Let $d$ denote the hidden size and $d_{\mathrm{KV}}$ the total key/value width across heads.
For a group $g$, let $h_r\in\mathbb{R}^{d}$ denote the normalized layer input of its $r$-th Summary, where $0\leq r<k_m$.
We compute its key, value, and aggregation logits as
\begin{equation}
 \begin{aligned}
  k_r&=\operatorname{RMSNorm}(W_K h_r),
  \qquad v_r=W_V h_r,\\
  z_r&=A h_r.
 \end{aligned}
\end{equation}
Here, $W_K,W_V,A\in\mathbb{R}^{d_{\mathrm{KV}}\times d}$; $W_K$ and $W_V$ are the backbone's existing projections, and $A$ is an additional learned projection.
RMSNorm is applied independently to each attention head.
We add a learned within-group position bias $B_r\in\mathbb{R}^{d_{\mathrm{KV}}}$ and normalize the logits across the $k_m$ Summaries to obtain weight vectors $\alpha_r\in\mathbb{R}^{d_{\mathrm{KV}}}$:
\begin{equation}
 [\alpha_0,\ldots,\alpha_{k_m-1}]
 =\operatorname{softmax}_{\mathrm{group}}
 \bigl([z_0+B_0,\ldots,z_{k_m-1}+B_{k_m-1}]\bigr).
\end{equation}
The softmax operates independently for each feature coordinate.
The same weights aggregate the keys and values:
\begin{equation}
 \bar{k}=\sum_{r=0}^{k_m-1}\alpha_r\odot k_r,
 \qquad
 \bar{v}=\sum_{r=0}^{k_m-1}\alpha_r\odot v_r,
 \label{eq:cksa-pooling}
\end{equation}
where $\odot$ denotes element-wise multiplication.
Finally, we apply rotary position embedding (RoPE)~\citep{rw_rope} to the pooled key at position $p_g$ of the group's last Summary, producing the Meta KV entry
\begin{equation}
 M_g=\bigl(\operatorname{RoPE}(\bar{k},p_g),\,\bar{v}\bigr).
\end{equation}
The aggregation parameters $A$ and $\{B_r\}_{r=0}^{k_m-1}$ are specific to each layer and shared across groups.
Meta entries are used only within the layer that constructs them, while the original Summary states continue to propagate across layers.
The group size $k_m$ can therefore vary across C-KSA layers without changing the shared Summary token layout.
We omit the layer index for clarity.

\paragraph{Visible KV entries.}
For each text query, C-KSA attends to three types of KV entries: local text, boundary Summaries, and remote Meta entries (Figure~\ref{fig:cksa-hierarchy}).
The local window contains the causal prefix of the current chunk and up to $W$ preceding complete chunks.
Outside this window, each complete group of $k_m$ Summaries is represented by one Meta KV entry.
The remaining fewer than $k_m$ Summaries immediately preceding the window remain individually accessible until their entire group lies outside the window.
All three types of entries participate in a single attention softmax.

\paragraph{Attention-mass correction.}
\fourbrevision{Compressing remote Summary entries can shift attention toward the local window by reducing the number of terms in the softmax denominator, even if local logits remain unchanged.
We therefore add $\log k_m$ to each Meta logit $s$ before normalization.
Since $\exp(s+\log k_m)=k_m\exp(s)$, this helps preserve the balance between local and remote attention.}


\subsection{Sparse KSA (S-KSA)}
\label{sec:method-sparse}

\begin{figure}[t]
  \centering
  \includegraphics[width=\linewidth]{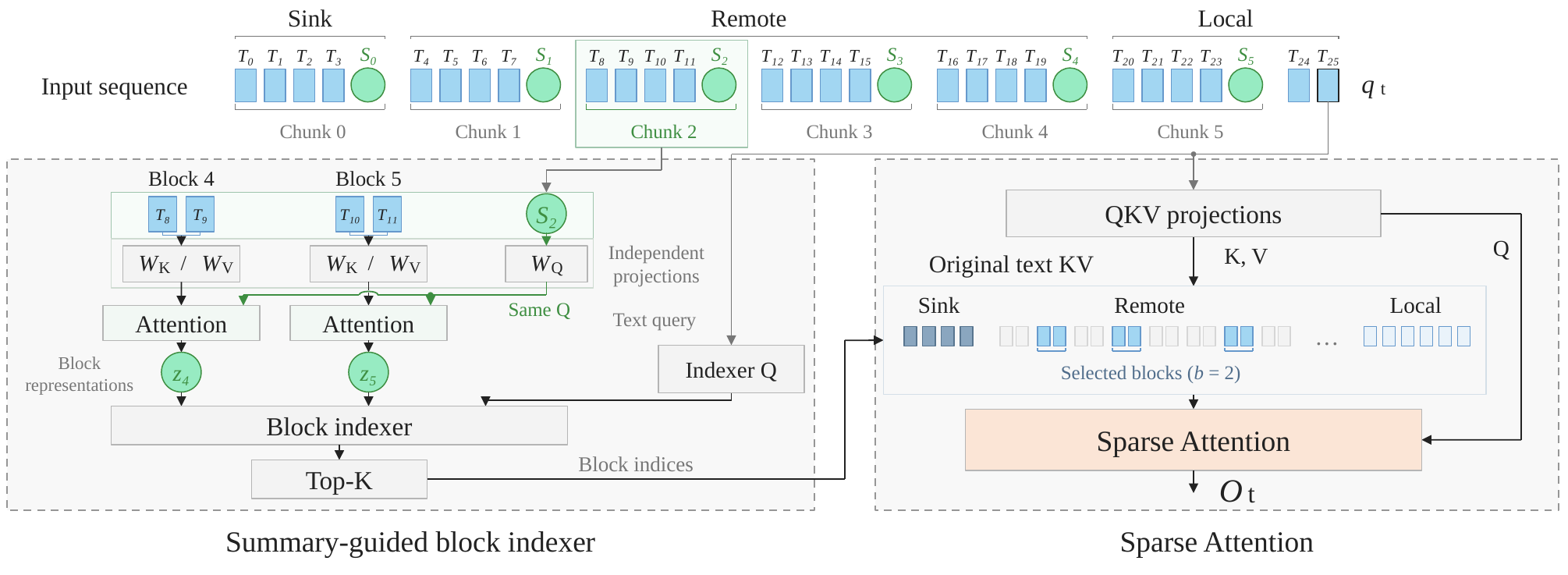}
  \caption{Sparse KSA with $k_s=4$ and $b=2$.}
  \label{fig:summary-block-sparse}
\end{figure}

S-KSA replaces full attention with Summary-guided block selection and sparse reads of original text (Figure~\ref{fig:summary-block-sparse}).
It uses the Summary tokens already present in KSA to construct one index key for each text block.
For each query, a Lightning Indexer selects relevant remote blocks, and the main attention reads their original KV together with local context.
This design reduces the routing overhead of token-level selection and avoids dense attention over the entire text history, while preserving direct access to the selected blocks' original tokens.

\paragraph{Summary-guided block aggregation.}
We divide each chunk into blocks of $b$ consecutive text tokens, where $b$ divides $k_s$.
For a block $u$, let $h_1,\ldots,h_b\in\mathbb{R}^{d}$ denote its normalized layer inputs and $h_S\in\mathbb{R}^{d}$ the normalized input of its chunk's Summary.
For one auxiliary attention head of width $d_A$, omitting the head index, we compute
\begin{equation}
 \begin{aligned}
  q_S&=\operatorname{RoPE}\!\left(
    \operatorname{RMSNorm}(W_Q^A h_S),p_S
  \right),\\
  k_i&=\operatorname{RoPE}\!\left(
    \operatorname{RMSNorm}(W_K^A h_i),p_i
  \right),\\
  v_i&=W_V^A h_i.
 \end{aligned}
\end{equation}
Here, $W_Q^A,W_K^A,W_V^A\in\mathbb{R}^{d_A\times d}$ are auxiliary projections specific to each S-KSA layer and independent of the backbone's attention projections.
The positions $p_S$ and $p_i$ follow the original sequence positions, with the Summary sharing its chunk's last text position.

Writing $K^A=[k_1,\ldots,k_b]$ and $V^A=[v_1,\ldots,v_b]$, the head's block representation is
\begin{equation}
 \widehat z_u
 =V^A\operatorname{softmax}\!\left(
   \frac{(K^A)^\top q_S}{\sqrt{d_A}}
 \right).
 \label{eq:kbsa-block-aggregation}
\end{equation}
The softmax operates over the block's $b$ text tokens; the Summary supplies only the query.
Concatenating the outputs of the $H_A$ auxiliary heads gives $z_u\in\mathbb{R}^{H_A d_A}$.
Blocks in a chunk share the Summary query but aggregate their own text tokens.
This resembles the landmark pooling of HiLS-Attention~\citep{rw_hils}, but reuses the existing Summary as the query rather than inserting additional landmark tokens.

\paragraph{Block indexing.}
We adapt the Lightning Indexer~\citep{rw_deepseekv32} to score text blocks using their Summary-guided representations.
For a text query at position $t$, the score for block $u$ is
\begin{equation}
 \rho_u=\sum_{h=1}^{H_I} w_h\,
   \operatorname{ReLU}\!\left(
       \frac{(q_h^I)^\top k_u^I}{\sqrt{d_I}}
   \right).
 \label{eq:indexer-score}
\end{equation}
The Indexer uses $H_I$ query heads of dimension $d_I$.
For each text query, its normalized layer input $h_t$ is mapped to a query vector $q_h^I\in\mathbb{R}^{d_I}$ and a scalar weight $w_h$ for each head $h$.
Each block representation $z_u$ is mapped to a single index key $k_u^I\in\mathbb{R}^{d_I}$, shared across all heads.
The weights $w_h$, scaled by $1/\sqrt{H_I}$, combine the per-head relevance scores into the block score $\rho_u$.
\fourbrevision{For the 4B model, we additionally apply RMSNorm to the projected index queries and keys before RoPE to improve adaptation convergence. Section~\ref{sec:scaling-4b} discusses this choice.}
Before scoring, partial RoPE encodes queries at position $t$ and keys at their blocks' last text positions.

\paragraph{Sparse access to original text.}
Given the block scores $\{\rho_u\}$, S-KSA selects up to $K/b$ remote blocks, where $K$ denotes the remote budget measured in original-text tokens.
Candidates must belong to completed chunks and lie entirely before the causal local window.
Eligible sink blocks at the beginning of the sequence are retained within this budget, and the remaining slots are filled by the highest-scoring candidates.
Let $\mathcal L_t$ and $\mathcal R_t$ denote local and selected remote positions, respectively.
The main attention computes
\begin{equation}
 O_t
 =\operatorname{Attn}\!\left(
   q_t,\,
   [K_{\mathcal L_t},K_{\mathcal R_t}],\,
   [V_{\mathcal L_t},V_{\mathcal R_t}]
 \right),
 \label{eq:kbsa-sparse-attention}
\end{equation}
where brackets denote concatenation along the sequence dimension.
The attention uses the backbone's original-text Q/K/V, with a single softmax over local and remote tokens in each head.

For $N$ text tokens, S-KSA constructs approximately $N/b$ index keys, reducing candidate comparisons by a factor of $b$ relative to token-level indexing.
Its routing cost scales as $O(N^2/b)$ during prefill, while the main attention reads a budgeted set of original-text blocks for each query.

\subsection{KV Cache for C-KSA}
\label{sec:kv-cache}

The C-KSA KV cache combines a contiguous attention region with a Summary ring buffer to support incremental decoding and second-level compression.
We first describe its layout and update rules, then analyze its memory cost and how sequence compression combines with GQA and MLA.

\subsubsection{Cache Structure and Updates}
\label{sec:method-cache}

To support efficient decoding, we design a compact KV cache structure for C-KSA (Figure~\ref{fig:cksa-cache}).
This structure allows each text query to read all visible KV entries through a single contiguous slice, avoiding per-step concatenation of cache segments.
Each C-KSA layer maintains two components: an \emph{Attention KV Cache} and a \emph{Summary Ring Buffer}.
The illustration uses $k_s=2$, $W=4$, and $k_m=4$, with $T_i$, $S_c$, and $M_g$ denoting text, Summary, and Meta KV entries, respectively.

\begin{figure}[t]
  \centering
  \input{figures/cache/plate}
  \caption{C-KSA cache layout and decoding updates with $k_s=2$, $W=4$, and $k_m=4$.
  Gray ($T$), orange ($S$), and purple ($M$) denote text, Summary, and Meta KV. Blue marks indicate reads or writes.}
  \label{fig:cksa-cache}
\end{figure}

\paragraph{Cache layout.}
The Attention KV Cache stores keys and values in separate tensors with the same layout: the current chunk, a local text ring, and remote history, followed by free slots (Figure~\ref{fig:cksa-cache-layout}).
The current chunk reserves $k_s$ text slots, and the local ring reserves $k_s W$ slots for $W$ preceding complete chunks.
Remote history contains Meta entries followed by fewer than $k_m$ Summaries awaiting compression.
The visible KV entries are contiguous along the sequence dimension, allowing each text query to read them with a single slicing operation.
The Summary Ring Buffer stores the local chunks' Summary KV and auxiliary compression state.
While a chunk remains in the local window, text queries attend to its original-text KV rather than its Summary KV.
When the chunk leaves the window, its Summary KV enters remote history.
A shared pointer identifies the next chunk slot to replace in both rings.

\paragraph{Text and Summary token insertion.}
New text KV entries are written from right to left in the current-chunk region, keeping the filled portion adjacent to the local text ring.
Each text query attends to the visible KV, including its own entry.
At a chunk boundary, the Summary query attends only to the completed chunk's text KV and its own KV.
In the example, $T_{23}$ completes chunk~11 (Figure~\ref{fig:cksa-cache-text}), and $S_{11}$ attends to $T_{22}$, $T_{23}$, and itself (Figure~\ref{fig:cksa-cache-summary}).

\paragraph{Updating the text and Summary rings.}
Once text and Summary attention finish, the completed chunk's text KV is copied into the local text ring, and its Summary KV and auxiliary compression state are stored in the Summary Ring Buffer.
If the rings are not yet full, the new entries occupy unused slots.
Otherwise, they replace the oldest local chunk and its Summary record; the old Summary KV moves into remote history, with its auxiliary state retained for compression.
The shared pointer advances to the next chunk slot.
In Figure~\ref{fig:cksa-cache-ring}, chunk~11 replaces chunk~7: $S_7$ enters remote history, $S_{11}$ takes its Summary ring slot, and the pointer advances to the slot holding $S_8$.

\paragraph{Compressing remote Summaries.}
When $k_m$ uncompressed Summaries accumulate in remote history, they are compressed into a single Meta KV entry.
This entry overwrites the group's first slot, leaving the remaining $k_m-1$ slots available for reuse.
In Figure~\ref{fig:cksa-cache-meta}, $S_4,S_5,S_6,S_7$ are compressed into $M_1$, stored immediately after $M_0$.
The visible KV entries remain contiguous, allowing each text query to read them through a single slice.

\subsubsection{KV Cache Memory Analysis}
\label{sec:memory-analysis}

We extend the KV-cache analysis of KSA with C-KSA's second compression level.
Table~\ref{tab:kv-cache} retains the original mechanisms for comparison.
Table~\ref{tab:kv-cache} counts stored scalars per layer and request, excluding allocation headroom and auxiliary buffers.

\begin{table}[t]
 \centering
 \small
 \setlength{\tabcolsep}{4pt}
 \renewcommand{\arraystretch}{1.3}
 \caption{KV-cache cost comparison for a sequence of $N$ text tokens.
 Example settings use $k_s=8$, $k_m=128$, $H=128$, $d_h=128$, $H_{\mathrm{KV}}=8$, $d_c=512$, and $d_r=64$.
 Compression rates are relative to MHA.}
 \label{tab:kv-cache}
 \begin{tabular}{lccc}
  \toprule
  \textbf{Mechanism} & \textbf{Effective Context} & \textbf{KV Cache Size ($N\rightarrow\infty$)} & \textbf{Compression Rate} \\
  \midrule
  MHA & $N$ (exactly) & $2\cdot N\cdot H\cdot d_h$ & --- \\
  \midrule
  \rowcolor{yellow!10}
  GQA & $N$ (exactly) & $2\cdot N\cdot H_{\mathrm{KV}}\cdot d_h$ & $\frac{H_{\mathrm{KV}}}{H}\approx6.25\%$ \\
  \rowcolor{purple!10}
  MLA & $N$ (exactly) & $N\cdot(d_c+d_r)$ & $\frac{d_c+d_r}{2Hd_h}\approx1.76\%$ \\
  \midrule
  GDN & $N$ (ambiguous) & $H\cdot d_h^2$ & $\frac{d_h}{2N}\approx0\%$ \\
  \rowcolor{blue!10}
  SWA & $w$ (exactly) & $2\cdot w\cdot H_{\mathrm{KV}}\cdot d_h$ & $\frac{w}{N}\cdot\frac{H_{\mathrm{KV}}}{H}\approx0\%$ \\
  \rowcolor{blue!10}
  KSA & $N$ (summarized) & $2\cdot\frac{N}{k_s}\cdot H\cdot d_h$ & $\frac{1}{k_s}\approx12.5\%$ \\
  \midrule
  KSA with GQA & $N$ (summarized) & $2\cdot\frac{N}{k_s}\cdot H_{\mathrm{KV}}\cdot d_h$ & $\frac{1}{k_s}\cdot\frac{H_{\mathrm{KV}}}{H}\approx0.78\%$ \\
  KSA with MLA & $N$ (summarized) & $\frac{N}{k_s}\cdot(d_c+d_r)$ & $\frac{1}{k_s}\cdot\frac{d_c+d_r}{2Hd_h}\approx0.22\%$ \\
  \midrule
  \rowcolor{blue!10}
  C-KSA & $N$ (summarized) & $2\cdot\frac{N}{k_s k_m}\cdot H\cdot d_h$ & $\frac{1}{k_s k_m}\approx0.098\%$ \\
  C-KSA with GQA & $N$ (summarized) & $2\cdot\frac{N}{k_s k_m}\cdot H_{\mathrm{KV}}\cdot d_h$ & $\frac{1}{k_s k_m}\cdot\frac{H_{\mathrm{KV}}}{H}\approx0.0061\%$ \\
  C-KSA with MLA & $N$ (summarized) & $\frac{N}{k_s k_m}\cdot(d_c+d_r)$ & $\frac{1}{k_s k_m}\cdot\frac{d_c+d_r}{2Hd_h}\approx0.0017\%$ \\
  \bottomrule
 \end{tabular}
\end{table}

\paragraph{KV Cache Cost.}
\label{sec:cksa-cost}

Let $N$ denote text length, $H$ and $H_{\mathrm{KV}}$ the query and KV head counts, and $d_h$ the head width.
In Table~\ref{tab:kv-cache}, $d_c$ and $d_r$ are MLA's latent and rotary-key widths, and $w$ is the local text window.

\begin{samepage}
KSA's leading cache term is $2NH_{\mathrm{KV}}d_h/k_s$.
Under the compression rule in Section~\ref{sec:method-cksa}, C-KSA replaces it with
\begin{equation}
 M_{\text{C-KSA}}
 =2H_{\mathrm{KV}}d_h\left[\frac{N}{k_s k_m}+O(k_s W+W+k_m)\right].
 \label{eq:layer-memory}
\end{equation}
\end{samepage}%
The bounded term accounts for local text, recent Summaries, and unfinished compression groups.
Thus C-KSA reduces the growing history term by a further factor of $k_m$ while preserving the local window.
With $k_s=8$ and $k_m=128$, the growing history term is $1/1024$ of uncompressed attention with the same KV width.
This nominal $1024\times$ compression applies to distant history; the local and auxiliary terms remain.
For fixed compression factors, cache growth remains linear in $N$.

\paragraph{Orthogonality of KV Cache Compression.}
\label{sec:orthogonal}

C-KSA preserves KSA's orthogonality to per-position compression: the additional $k_m$ factor multiplies the sequence compression without changing the KV head count or representation width.
The resulting GQA and MLA combinations are listed in Table~\ref{tab:kv-cache}.

\begin{figure}[t]
 \centering
 \includegraphics[width=\linewidth]{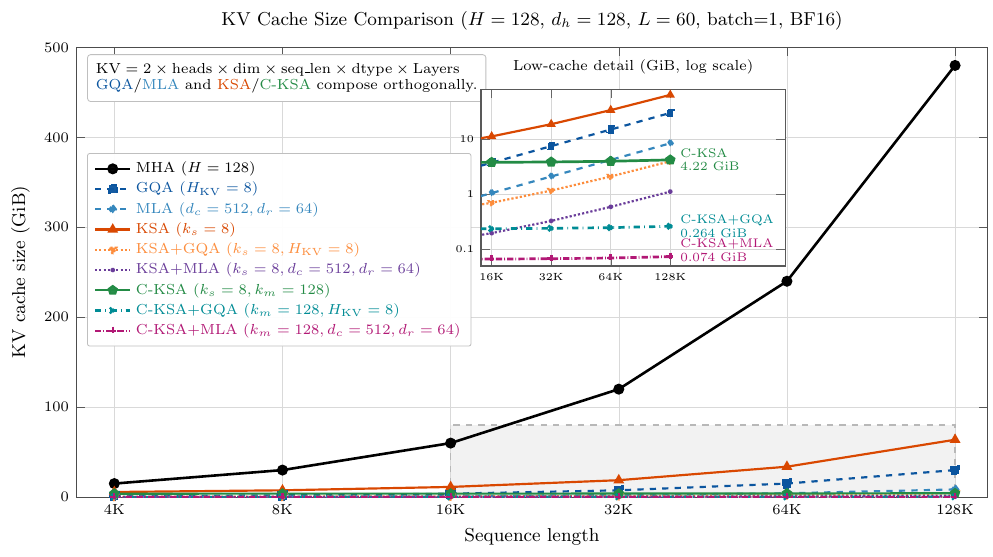}
 \caption{KV cache size comparison across different mechanisms as sequence length grows.
 The inset uses a logarithmic vertical axis to separate the low-cache curves and labels C-KSA values at 128K.}
 \label{fig:kv-cache-comparison}
\end{figure}

Figure~\ref{fig:kv-cache-comparison} uses the settings in Table~\ref{tab:kv-cache} with 60 layers, batch size 1, BF16, and $w=k_s W=1024$ local tokens.
The KSA and C-KSA curves use $N/k_s+w$ and $N/(k_s k_m)+w$ entries, respectively, omitting other bounded auxiliary buffers and allocation headroom.

\paragraph{Hybrid KSA and C-KSA.}
With matched layer counts and KV widths and a fraction $f$ of Full layers, the leading cache ratio is
\begin{equation}
 \frac{M_{\text{C-KSA+Full}}}{M_{\mathrm{Hybrid\ KSA}}}
 \simeq\frac{f+(1-f)/(k_s k_m)}{f+(1-f)/k_s}.
 \label{eq:hybrid-memory}
\end{equation}
With $f=1/4$ in the $3{:}1$ layout, the Full layers limit the whole-model savings.

\FloatBarrier
\subsection{Kernel Design}
\label{sec:kernel-design}

\begin{figure}[htbp]
  \centering
  \input{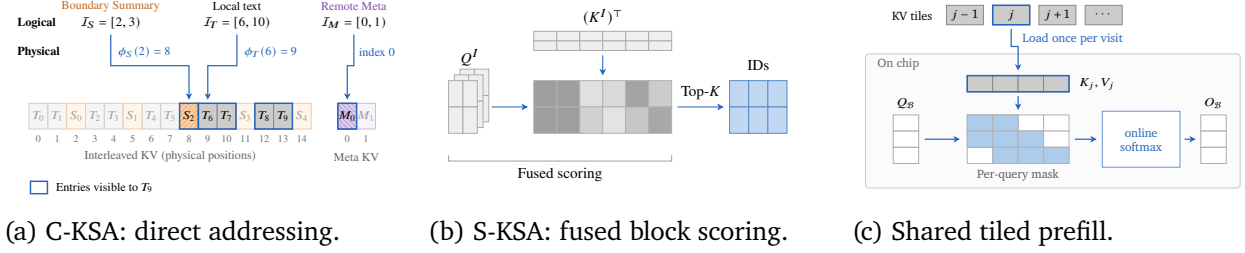}
  \caption{Kernel design for C-KSA and S-KSA: (a) direct KV addressing,
  (b) fused block scoring and selection under a $K$-token budget, and
  (c) shared tiled prefill with KV reuse.}
  \label{fig:ksa-kernel}
\end{figure}

C-KSA and S-KSA reduce attention computation and KV reads through compression and sparse access.
Realizing these savings on GPUs requires efficient KV addressing, block selection, and attention execution.
Figure~\ref{fig:ksa-kernel} summarizes the method-specific addressing and routing alongside the tiled prefill pattern used by both variants.

\paragraph{Shared prefill execution.}
Both C-KSA and S-KSA follow FlashAttention's tiled execution and online softmax~\citep{rw_flashattention}, as illustrated in Figure~\ref{fig:kernel-prefill}.
Let $\mathcal B$ denote a group of text queries arranged into one or more query tiles, with query and output matrices $Q_{\mathcal B}$ and $O_{\mathcal B}$.
Each box in these matrices represents one token vector; blue cells in the query--key grid indicate visible pairs, while white cells are masked.
Each loaded KV tile $(K_j,V_j)$ is reused on chip within the group, while a per-query mask enforces each query's visible set.
Building on this shared execution pattern, C-KSA and S-KSA specialize the KV addressing and tile schedule described below.

\subsubsection{C-KSA}
\label{sec:kernel-cksa}

\paragraph{Logical intervals and direct addressing.}
C-KSA's regular layout allows the kernel to compute the KV locations needed by each query using simple arithmetic.
During prefill, text and Summary Q/K/V are interleaved, while precomputed Meta K/V are stored in a separate buffer.
We treat text, Summary, and Meta entries as three logical segments, each with its own zero-based index space.

\begin{samepage}
Figure~\ref{fig:cksa-kernel-layout} illustrates this addressing process with $k_s=2$, $k_m=2$, and $W=1$. Blue outlines mark entries visible to $T_9$.
For a text query at position $t$, let $a=\max(0,\lfloor t/k_s\rfloor-W)$ denote the first chunk in its local window and $g=\lfloor a/k_m\rfloor$ the number of complete Meta groups preceding that window.
We first determine the visible logical indices in each segment:
\begin{equation}
 \mathcal I_T(t)=[k_s a,\,t+1),
 \qquad
 \mathcal I_S(t)=[k_m g,\,a),
 \qquad
 \mathcal I_M(t)=[0,\,g).
 \label{eq:cksa-kernel-ranges}
\end{equation}
\end{samepage}%
These intervals contain the causal local text, boundary Summaries, and complete remote Meta groups, respectively.

\begin{samepage}
We then map these logical indices to physical positions in the interleaved storage.
For text index $i$ and Summary index $j$, the mappings are
\begin{equation}
 \phi_T(i)=i+\left\lfloor\frac{i}{k_s}\right\rfloor,
 \qquad
 \phi_S(j)=j(k_s+1)+k_s.
 \label{eq:cksa-kernel-addresses}
\end{equation}
\end{samepage}%
The kernel applies these mappings when loading Q/K/V.
Meta indices directly address entries in the separate Meta buffer.
For each query tile, the logical intervals determine the KV tile ranges to visit, and a position-based mask excludes entries outside each query's visible set.
The ranges and masks are computed within the kernel, with no dense attention mask or block-index table stored in memory.

\paragraph{Prefill integration.}
The text-query kernel carries each query's online-softmax state across the text, Summary, and Meta segments, applying the $\log k_m$ correction only to Meta logits.
A separate local kernel computes each Summary query over its own text chunk and self-KV.
Both kernels write outputs directly to their interleaved positions.

\paragraph{Decoding.}
The KV cache in Section~\ref{sec:method-cache} stores the visible KV entries in a contiguous region, allowing attention to be computed directly with scaled dot-product attention (SDPA).

\subsubsection{S-KSA}
\label{sec:kernel-kbsa}

S-KSA uses fused block scoring and a sparse KV-tile schedule to support query-dependent access to original text.
The scoring optimization also applies during decoding.

\paragraph{Fused block scoring.}
Using separate kernels for dot products and head aggregation requires storing intermediate scores in GPU memory.
To reduce this memory traffic, we compute block scores with a fused kernel that performs dot products and combines weighted head scores on chip.
In Figure~\ref{fig:kbsa-kernel-selection}, $Q^I$ denotes the stacked query matrices of the indexer heads and $K^I$ their shared block keys. The superscript $I$ denotes the indexer.
Each $Q^I$ matrix has queries along rows and features along columns. The central grayscale matrix has queries along rows and candidate blocks along columns, with darker cells indicating higher scores.
A separate selection step then chooses remote blocks independently for each query under a $K$-token budget (up to $K/b$ blocks); each row of the blue IDs matrix on the right lists the block indices selected for one query.

\paragraph{KV-tile scheduling.}
For each query group, we map its selected remote blocks and local windows to KV tiles and construct a conservative schedule of tiles to visit.
The schedule skips inactive KV tiles, while per-query masks enforce block selection, local visibility, and causality within the scheduled tiles.
We supply this schedule and the masks to the FlashAttention forward kernel for tiled prefill execution.

\paragraph{Decoding.}
Gathering selected KV entries into temporary buffers incurs extra memory reads and writes before attention computation.
To avoid this overhead during decoding, we use a custom sparse kernel that reads directly from the KV cache using the selected block indices.

\FloatBarrier

\section{Experiments}
\label{sec:experiments}

Our experiments examine whether continued pretraining can adapt Hybrid KSA to more aggressive KV compression and sparse original-text access while retaining its capabilities.
We first evaluate C-KSA with Full attention, then add S-KSA to the compressed backbone, assessing long-context retrieval, general capabilities, and inference efficiency.
We further analyze the effects of compression granularity, block size, teacher pooling, and routing on model performance.
\fourbrevision{Unless otherwise stated, all experiments and analyses use the 1.9B model.}

\subsection{Experimental Setup}
\label{sec:experimental-setup}

\paragraph{Pretrained model and variants.}
\fourbrevision{We start from the 128K checkpoints of the 1.9B and 4B Hybrid KSA models in the KSA report~\citep{rw_ksa}.
The 1.9B backbone has 24 multi-head attention (MHA) layers; the 4B backbone has 36 grouped-query attention (GQA) layers with 32 query heads and eight KV heads.
Both interleave three KSA layers with one Full layer and insert one Summary after every $k_s=8$ text tokens.
We evaluate two successive variants: \textbf{C-KSA+Full} compresses remote Summary KV in KSA layers; \textbf{C-KSA+S-KSA} further replaces the Full layers with S-KSA.}
Although C-KSA supports layer-specific compression, each evaluated configuration uses a common $k_m$ across its C-KSA layers.
Unless otherwise specified, all experiments use the backbone and module configurations in Table~\ref{tab:model-configuration}.

\paragraph{Baselines and training data.}
\fourbrevision{We reuse the KSA report's from-scratch results at 1.9B and continual-pretraining results at 4B.
The baselines include Full, Hybrid SWA, Hybrid SCA, Pure KSA, and Hybrid KSA at both scales, plus Hybrid GDN at 1.9B and Hybrid Linear at 4B.}
Hybrid KSA serves as the parent-model reference for capability retention, while Full provides a dense-attention reference.
\fourbrevision{We report module-only and full-parameter training budgets separately.}
A continuation study found no consistent RULER improvement for \fourbrevision{1.9B} Hybrid KSA under the tested recipe and budgets (Table~\ref{tab:hybrid-ksa-continuation}), supporting our use of its original checkpoint as the capability reference.
\fourbrevision{Adaptation reuses KSA's training corpus and data mixture, with data sources and mixture proportions detailed in the original report.
We use a fixed context length of 128K.}

\paragraph{Evaluation.}
Following the KSA report, we assess long-context modeling and general capabilities.
Long-context modeling is evaluated using RULER~\citep{rw_ruler} at sequence lengths from 4K to 128K.
General capabilities cover knowledge (MMLU, CMMLU, C-Eval, and MMLU-Pro~\citep{hendrycksmeasuring,li2024cmmlu,huang2023c,wang2024mmlu}), mathematics (GSM8K, CMATH, and MATH~\citep{cobbe2021training,wei2023cmath,hendrycksmath2021}), and code generation (MBPP and HumanEval~\citep{austin2021program,chen2021evaluating}).
\begin{table}[t]
  \centering
  \caption{\fourbrevision{Configurations of the selected 1.9B and 4B models. Module settings apply where the corresponding modules are used.}}
  \label{tab:model-configuration}
  \small
  \renewcommand{\arraystretch}{1.15}
  \begin{tabular*}{\linewidth}{@{\extracolsep{\fill}}lrr@{}}
    \toprule
    \textbf{Configuration} & \fourbrevision{\textbf{1.9B}} & \textbf{4B} \\
    \midrule
    \multicolumn{3}{l}{\textbf{Shared backbone}} \\
    Number of layers & 24 & 36 \\
    Hidden size & 2048 & 2560 \\
    Intermediate size & 6144 & 9728 \\
    Attention heads (Q/KV) & 16/16 & 32/8 \\
    Head dimension & 128 & 128 \\
    Hybrid architecture ratio & $3{:}1$ & $3{:}1$ \\
    Summary chunk size $k_s$ & 8 & 8 \\
    Sliding chunk number (KSA/C-KSA) & 128 & 128 \\
    Tied embeddings & False & True \\
    \midrule
    \multicolumn{3}{l}{\textbf{C-KSA}} \\
    Summary-to-Meta group size $k_m$ & 128 & 32 \\
    KV width $d_{\mathrm{KV}}$ & 2048 & 1024 \\
    Aggregation projection $A$ shape & $2048\times2048$ & $1024\times2560$ \\
    Position-bias shape $k_m\times d_{\mathrm{KV}}$ & $128\times2048$ & $32\times1024$ \\
    Additional parameters & 80.2M & 71.7M \\
    \midrule
    \multicolumn{3}{l}{\textbf{S-KSA}} \\
    Block size $b$ & 4 & 4 \\
    Remote token budget $K$ & 4096 & 4096 \\
    Sink tokens, included in $K$ & 8 & 8 \\
    Local window length (text tokens) & 1024 & 1024 \\
    \fourbrevision{Aggregation heads (Q/KV)} & \fourbrevision{16/16} & 32/8 \\
    Aggregation head dimension $d_A$ & 128 & 128 \\
    Indexer heads & 8 & 8 \\
    Indexer head dimension & 128 & 128 \\
    Query low-rank dimension & 512 & 512 \\
    \fourbrevision{Final Indexer Q/K RMSNorm} & \fourbrevision{No} & Yes \\
    Additional parameters & 86.6M & 163.0M \\
    \bottomrule
  \end{tabular*}
  \par
\end{table}

\subsection{Adaptation Protocol}
\label{sec:training-protocol}

\begin{figure}[!htbp]
 \centering
 \includegraphics[width=\linewidth]{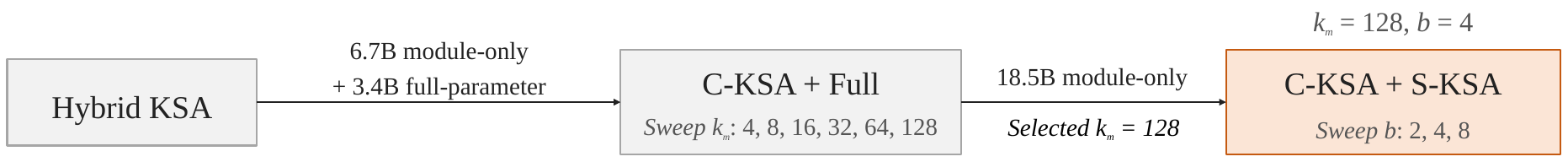}
 \caption{Sequential adaptation and parameter selection for KSA V2 \fourbrevision{at 1.9B}.}
 \label{fig:adaptation-routes}
\end{figure}

Figure~\ref{fig:adaptation-routes} summarizes our sequential adaptation route \fourbrevision{at 1.9B}: we first study compression granularity in C-KSA+Full, then compare sparse-attention schemes on the selected $k_m=128$ checkpoint.
Module-only adaptation trains only newly introduced parameters while freezing all inherited parameters; full-parameter adaptation updates the entire model.
C-KSA uses module-only adaptation followed by full-parameter adaptation; the subsequent S-KSA stage uses module-only adaptation.

\paragraph{C-KSA adaptation.}
During module-only adaptation, we train the aggregation projection $A$ and position biases $\{B_r\}$ while freezing all inherited parameters.
The new parameters are initialized to zero, yielding uniform aggregation weights.
We align compressed attention outputs with those of uncompressed KSA.
Let $O^{\mathrm C}_{l,t,d}$ and $O^{\mathrm K}_{l,t,d}$ denote the $d$-th coordinate of the attention outputs from C-KSA and uncompressed KSA, respectively, at layer $l$ and text position $t$.
Both outputs are computed from the same layer input, before the output projection.
The alignment loss is
\begin{equation}
 \mathcal L_{\mathrm{align}}
 =
 \frac{1}{|\mathcal I_{\mathrm C}|\,|\mathcal T|\,D}
 \sum_{l\in\mathcal I_{\mathrm C}}
 \sum_{t\in\mathcal T}
 \sum_{d=1}^{D}
 \min\!\left\{
   \left(O^{\mathrm C}_{l,t,d}
   -\operatorname{sg}(O^{\mathrm K}_{l,t,d})\right)^2,\;0.1
 \right\},
 \label{eq:compression-alignment}
\end{equation}
where $\mathcal I_{\mathrm C}$ is the set of C-KSA layers, $\mathcal T$ contains the text positions in a minibatch, $D$ is the concatenated attention width, and $\operatorname{sg}(\cdot)$ denotes stop-gradient.
Each squared coordinate error is clipped before averaging.
We then transition from module-only to full-parameter adaptation, using the objectives
\begin{equation}
 \mathcal L_{\mathrm{comp}}
 =
 \begin{cases}
  \mathcal L_{\mathrm{CE}}+\lambda_{\mathrm{align}}\mathcal L_{\mathrm{align}},
    & \text{module-only adaptation},\\[3pt]
  \mathcal L_{\mathrm{CE}},
    & \text{full-parameter adaptation},
 \end{cases}
 \label{eq:compression-loss}
\end{equation}
where $\mathcal L_{\mathrm{CE}}$ is the text-token cross-entropy loss and $\lambda_{\mathrm{align}}=5$.

\paragraph{S-KSA adaptation.}
Module-only adaptation trains the new aggregation modules and Indexer while freezing all inherited parameters, including the C-KSA aggregation parameters.
During training, we retain dense attention in Full layers and use its probabilities to supervise block selection.
For query $t$, let $\mathcal U_t$ contain all eligible remote blocks, $\mathcal B_u$ the text positions in block $u$, and $a_{t,i}$ the dense causal attention probability averaged across backbone heads.
Suppressing the layer index, we define the block-level teacher distribution $p_t$, Indexer distribution $\pi_t$, and training objective as
\begin{equation}
 \begin{aligned}
  p_{t,u}
  &=
  \operatorname{sg}\!\left(
   \frac{\max_{i\in\mathcal B_u}a_{t,i}}
        {\sum_{v\in\mathcal U_t}\max_{j\in\mathcal B_v}a_{t,j}}
  \right),\\[4pt]
  \pi_t
  &=
  \operatorname{softmax}(\rho_t),\\[4pt]
  \mathcal L_{\text{S-KSA}}
  &=
  \mathbb E_{l,t}\!\left[
   D_{\mathrm{KL}}\!\left(p_t\,\|\,\pi_t\right)
  \right],
 \end{aligned}
 \label{eq:indexer-loss}
\end{equation}
where $\operatorname{sg}(\cdot)$ denotes stop-gradient and $\rho_t$ contains the Indexer scores defined in Equation~\ref{eq:indexer-score} for all blocks in $\mathcal U_t$.
The loss is averaged over valid text queries within each adapted layer, then across layers.
Following QSA~\citep{rw_qsa}, the teacher takes the blockwise maximum, after averaging over heads; we compare max and sum pooling in Section~\ref{sec:ablation-studies}.
At inference, the trained Indexer selects remote blocks for sparse attention.

\paragraph{Training configuration.}
Table~\ref{tab:adaptation-recipe} summarizes our adaptation hyperparameters.
We reinitialize the optimizer and learning-rate schedule at each stage transition.

\begin{table}[t]
  \centering
  \caption{Adaptation hyperparameters and token budgets for the selected models.}
  \label{tab:adaptation-recipe}
  \small
  \renewcommand{\arraystretch}{1.15}
  \setlength{\tabcolsep}{5pt}
  \begin{tabular}{lccc}
    \toprule
    \multirow{2}{*}{\textbf{Configuration}} & \multicolumn{2}{c}{\textbf{C-KSA}} & \textbf{S-KSA} \\
    \cmidrule(lr){2-3}\cmidrule(lr){4-4}
    & Module-only & Full-parameter & Module-only \\
    \midrule
    Trainable parameters & Aggregation modules & All & Aggregation and Indexer \\
    Training tokens (B), \fourbrevision{1.9B} & 6.7 & 3.4 & 18.5 \\
    \fourbrevision{Training tokens (B), 4B} & \fourbrevision{6.7} & \fourbrevision{15.9} & \fourbrevision{4.2} \\
    Target learning rate & $10^{-4}$ & $10^{-4}$ & $10^{-3}$ \\
    LR warmup (updates) & 200 & 400 & 200 \\
    \midrule
    Sequence length & \multicolumn{3}{c}{128K} \\
    Global batch size & \multicolumn{3}{c}{32} \\
    Initial learning rate & \multicolumn{3}{c}{$10^{-6}$} \\
    LR schedule & \multicolumn{3}{c}{Warmup then constant} \\
    Optimizer & \multicolumn{3}{c}{AdamW} \\
    AdamW $(\beta_1,\beta_2)$ & \multicolumn{3}{c}{$(0.9,0.95)$} \\
    Weight decay & \multicolumn{3}{c}{0.01} \\
    Gradient clipping & \multicolumn{3}{c}{1.0} \\
    \bottomrule
  \end{tabular}
\end{table}

\subsection{Model Quality}
\label{sec:model-quality}

\begin{table}[t]
\centering
\footnotesize
\setlength{\tabcolsep}{1.5pt}
\caption{Long-context retrieval and general capabilities \fourbrevision{at 1.9B} (\%; pass@1 for code). Bold and underlined values mark the best and second-best scores, respectively, here and in subsequent tables.}
\label{tab:main-comparison}
\begin{tabular*}{\linewidth}{@{\extracolsep{\fill}}lcccccccc@{}}
\toprule
 & \multicolumn{6}{c}{Baselines} & \multicolumn{2}{c}{Ours} \\
\cmidrule(lr){2-7}\cmidrule(lr){8-9}
Benchmark & Full & Hybrid SWA & Hybrid SCA & Hybrid GDN & Pure KSA & Hybrid KSA & C-KSA+Full & C-KSA+S-KSA \\
\midrule
\rowcolor{gray!10}[0pt][0pt]
\multicolumn{9}{@{}l@{}}{\textbf{Long-context retrieval}} \\
RULER-4K & 76.08 & 74.54 & 77.72 & 79.83 & 70.44 & 80.65 & \textbf{81.52} & \underline{81.48} \\
RULER-8K & 72.85 & 71.69 & \underline{75.22} & \textbf{76.01} & 65.91 & 73.35 & 72.24 & 72.42 \\
RULER-16K & 73.24 & 69.54 & 72.55 & 74.04 & 66.74 & 74.07 & \underline{74.44} & \textbf{75.67} \\
RULER-32K & 69.06 & 67.86 & 67.74 & 70.41 & 62.54 & \textbf{72.30} & \underline{71.93} & 70.78 \\
RULER-64K & 65.32 & 63.03 & 63.54 & 69.39 & 57.13 & \textbf{69.95} & \underline{69.61} & 68.99 \\
RULER-128K & 48.75 & 56.64 & 58.01 & 59.87 & 39.29 & \underline{65.35} & \textbf{66.31} & 65.01 \\
RULER mean & 67.55 & 67.22 & 69.13 & 71.59 & 60.34 & \underline{72.61} & \textbf{72.68} & 72.39 \\
\midrule
\rowcolor{gray!10}[0pt][0pt]
\multicolumn{9}{@{}l@{}}{\textbf{General knowledge}} \\
MMLU & 44.99 & 46.84 & 46.77 & 46.23 & 46.83 & 46.83 & \underline{46.97} & \textbf{47.09} \\
CMMLU & 44.41 & 45.89 & 46.42 & \textbf{47.19} & 45.59 & \underline{46.88} & 46.49 & 46.39 \\
C-Eval & 44.28 & 43.54 & \textbf{47.62} & 45.54 & \underline{45.69} & 44.13 & 43.83 & 42.94 \\
MMLU-Pro & 19.48 & 20.46 & 20.10 & 21.22 & 21.72 & 22.52 & \underline{22.71} & \textbf{22.83} \\
\midrule
\rowcolor{gray!10}[0pt][0pt]
\multicolumn{9}{@{}l@{}}{\textbf{Mathematics}} \\
GSM8K & 48.29 & 47.46 & 52.39 & 50.95 & 54.81 & \textbf{59.14} & 55.95 & \underline{56.33} \\
CMATH & 55.33 & 54.83 & 62.33 & 58.00 & 58.50 & 61.83 & \textbf{64.50} & \underline{63.50} \\
MATH & 23.38 & 31.46 & 28.82 & 33.30 & 30.04 & \textbf{36.92} & 36.60 & \underline{36.72} \\
\midrule
\rowcolor{gray!10}[0pt][0pt]
\multicolumn{9}{@{}l@{}}{\textbf{Code}} \\
MBPP & 30.60 & 30.00 & 31.60 & 34.80 & \underline{35.80} & \textbf{36.40} & 35.60 & 35.20 \\
HumanEval & 25.61 & 28.05 & 26.83 & 27.44 & \underline{29.88} & \textbf{31.71} & 28.05 & \underline{29.88} \\
\bottomrule
\end{tabular*}
\end{table}

Table~\ref{tab:main-comparison} compares C-KSA+Full and C-KSA+S-KSA with the \fourbrevision{1.9B} baselines from the KSA report.

\paragraph{Long-context retrieval.}
C-KSA+S-KSA achieves a six-length RULER mean of 72.39\% and a 128K score of 65.01\%, compared with 72.61\% and 65.35\% for Hybrid KSA and 67.55\% and 48.75\% for Full.
The combined model thus recovers near-parent retrieval performance while retaining the advantage of Hybrid KSA over Full.

\paragraph{Compression and sparsity compose.}
At $k_m=128$, C-KSA+Full increases the distant-history compression factor within KSA layers from 8 to 1024, with mean and 128K RULER scores of 72.68\% and 66.31\%.
Adding S-KSA to this backbone lowers the 128K score to 65.01\%, whereas the same sparse adaptation applied to Hybrid KSA lowers it from 65.35\% to 61.06\% (Table~\ref{tab:architectural-ablations}(c)).

\paragraph{General capabilities.}
C-KSA+S-KSA remains broadly comparable to Hybrid KSA across the nine general-capability benchmarks, with gains on MMLU, MMLU-Pro, and CMATH and lower scores on the remaining tasks.
The largest decrease is 2.81 percentage points on GSM8K, while MBPP and HumanEval decrease by 1.20 and 1.83 points, respectively.
Compared with C-KSA+Full, adding S-KSA changes each score by at most 1.83 points, with both gains and losses, indicating no uniform decline from sparse adaptation.
The combined model exceeds Full on eight of the nine benchmarks, with C-Eval as the exception.

\begin{figure}[!htbp]
 \centering
 \includegraphics[width=\linewidth]{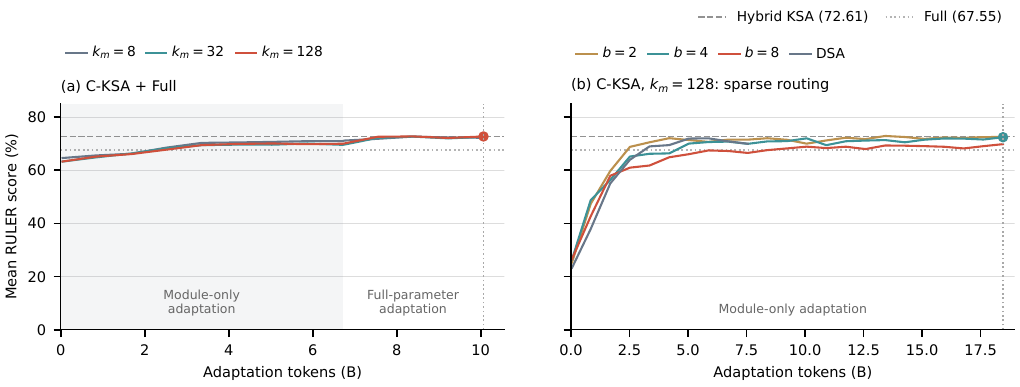}
 \caption{Mean RULER scores during compression and sparse adaptation. Circles mark the checkpoints reported in Table~\ref{tab:main-comparison}.}
 \label{fig:adaptation-dynamics}
\end{figure}

\paragraph{Adaptation dynamics.}
Figure~\ref{fig:adaptation-dynamics} shows how retrieval performance recovers during adaptation.
For C-KSA, retrieval performance improves during module-only adaptation and approaches Hybrid KSA after full-parameter adaptation.
On the selected C-KSA+Full backbone, S-KSA with $b=4$ approaches Hybrid KSA's performance by training only its newly introduced aggregation modules and Indexer.

\FloatBarrier

\subsection{\fourbrevision{Scaling to a 4B GQA Model}}
\label{sec:scaling-4b}

\fourbrevision{To test transfer beyond the 1.9B MHA backbone, we adapt the 4B GQA model with $k_m=32$ and $b=4$.
With $k_s=8$, C-KSA compresses distant-history KV by $256\times$ along the sequence dimension while preserving GQA's shared KV heads.
The C-KSA+Full checkpoint uses 6.7B tokens of module-only adaptation followed by 15.9B tokens of full-parameter adaptation.
Starting from this checkpoint, S-KSA uses a further 4.2B tokens of module-only adaptation, leaving all inherited parameters frozen.
The remote budget, local window, and sink settings remain unchanged from the 1.9B experiments; the 4B Indexer additionally normalizes its projected queries and keys as described in Section~\ref{sec:method-sparse}.}

\begin{table}[t]
\centering
\footnotesize
\setlength{\tabcolsep}{1.5pt}
\caption{\fourbrevision{Long-context retrieval and general capabilities at 4B (\%; pass@1 for code). Baselines are from the KSA report~\citep{rw_ksa}.}}
\label{tab:scaling-4b}
\begin{tabular*}{\linewidth}{@{\extracolsep{\fill}}l*{8}{c}@{}}
\toprule
 & \multicolumn{6}{c}{Baselines} & \multicolumn{2}{c}{Ours} \\
\cmidrule(lr){2-7}\cmidrule(lr){8-9}
Benchmark & \scriptsize Full & \scriptsize Hybrid SWA & \scriptsize Hybrid SCA & \scriptsize Hybrid Linear & \scriptsize Pure KSA & \scriptsize Hybrid KSA & \scriptsize C-KSA+Full & \scriptsize C-KSA+S-KSA \\
\midrule
\rowcolor{gray!10}[0pt][0pt]
\multicolumn{9}{@{}l@{}}{\textbf{Long-context retrieval}} \\
RULER-4K & 92.88 & 91.30 & 86.02 & 86.39 & 91.55 & \underline{92.97} & 92.77 & \textbf{93.38} \\
RULER-8K & \textbf{91.38} & 88.03 & 84.28 & 83.86 & 86.78 & \underline{90.53} & 88.12 & 87.78 \\
RULER-16K & \textbf{89.12} & 82.87 & 80.67 & 78.06 & 84.78 & \underline{88.86} & 86.27 & 85.45 \\
RULER-32K & \underline{84.74} & 78.94 & 76.89 & 76.48 & 80.30 & \textbf{86.65} & 82.68 & 83.60 \\
RULER-64K & 78.16 & 73.88 & 68.88 & 73.50 & 76.09 & 76.04 & \underline{78.58} & \textbf{80.28} \\
RULER-128K & 65.86 & 66.27 & 60.94 & 67.98 & 66.81 & 71.67 & \underline{73.92} & \textbf{74.41} \\
RULER mean & 83.69 & 80.22 & 76.28 & 77.71 & 81.05 & \textbf{84.45} & 83.72 & \underline{84.15} \\
\midrule
\rowcolor{gray!10}[0pt][0pt]
\multicolumn{9}{@{}l@{}}{\textbf{General knowledge}} \\
MMLU & \textbf{71.83} & 70.57 & 69.83 & 64.33 & \underline{70.73} & 70.50 & 68.79 & 68.79 \\
CMMLU & \textbf{75.00} & \underline{73.69} & 72.59 & 68.41 & 73.29 & 72.63 & 70.88 & 71.06 \\
C-Eval & \textbf{73.66} & 72.36 & 71.66 & 67.42 & 72.14 & \underline{72.66} & 70.36 & 70.58 \\
MMLU-Pro & \textbf{46.36} & 45.23 & 45.11 & 38.83 & \underline{45.70} & 45.39 & 43.66 & 43.58 \\
\midrule
\rowcolor{gray!10}[0pt][0pt]
\multicolumn{9}{@{}l@{}}{\textbf{Mathematics}} \\
GSM8K & \textbf{82.75} & \underline{81.92} & 80.10 & 72.44 & 81.09 & 79.50 & 79.00 & 79.00 \\
CMATH & 83.41 & \textbf{84.84} & 83.16 & 79.09 & \underline{84.58} & 84.25 & 81.50 & 81.67 \\
MATH & 47.48 & \textbf{48.24} & 47.45 & 42.57 & \underline{48.15} & 47.56 & 46.10 & 46.40 \\
\midrule
\rowcolor{gray!10}[0pt][0pt]
\multicolumn{9}{@{}l@{}}{\textbf{Code}} \\
MBPP & 61.30 & \underline{61.70} & 59.60 & 55.30 & 61.50 & \textbf{62.20} & 60.40 & 60.00 \\
HumanEval & 58.54 & \underline{61.89} & \underline{61.89} & 54.58 & 60.97 & \textbf{62.50} & 60.98 & 60.98 \\
\bottomrule
\end{tabular*}
\end{table}

\fourbrevision{Table~\ref{tab:scaling-4b} compares the adapted models with the six 4B baselines reported in KSA.
C-KSA+S-KSA achieves a six-length RULER mean of 84.15\%, compared with 84.45\% for Hybrid KSA and 83.72\% for its C-KSA+Full starting point.
At 128K, it scores 74.41\%, compared with 71.67\% for Hybrid KSA and 65.86\% for Full.
On general-capability benchmarks, C-KSA+S-KSA stays within 0.40 percentage points of C-KSA+Full and 0.50--2.58 points below Hybrid KSA.
Overall, C-KSA+S-KSA retains retrieval performance close to Hybrid KSA, with modest declines in general capabilities.
These results support its effectiveness at a larger scale and compatibility with GQA.}

\paragraph{\fourbrevision{Indexer Q/K normalization.}}
\fourbrevision{Without Q/K normalization, the 4B Indexer loss remains high, and RULER performance stays below its C-KSA+Full parent even after 21.8B adaptation tokens.
Under the same adaptation setup, normalizing the projected queries and keys restores near-parent performance with only 4.2B tokens.
A possible explanation is that differences in feature scale between the pretrained backbones affect index-logit magnitudes and optimization; RMSNorm reduces this scale sensitivity.
We retain the original 1.9B setup for consistency and have not repeated its full experimental suite with this change.
Whether normalization also reduces the adaptation data required at 1.9B remains an open question for future work.}

\paragraph{\fourbrevision{Compression granularity.}}
\fourbrevision{Table~\ref{tab:compression-granularity-4b} compares compression granularities on the 4B backbone, using 6.7B tokens of module-only adaptation followed by 15.9B tokens of full-parameter adaptation for each configuration.
Compared with $k_m=8$, $k_m=32$ reduces the number of remote Meta entries by a further factor of four while retaining similar RULER performance.
We select $k_m=32$ for the 4B sparse-adaptation experiments.}

\begin{table}[!htbp]
\centering
\small
\caption{\fourbrevision{RULER scores (\%) of 4B C-KSA+Full with different compression granularities.}}
\label{tab:compression-granularity-4b}
\begin{tabular*}{\linewidth}{@{\extracolsep{\fill}}l*{7}{r}@{}}
\toprule
$k_m$ & 4K & 8K & 16K & 32K & 64K & 128K & Mean \\
\midrule
8 & 91.13 & 88.22 & 85.71 & 82.62 & \textbf{81.35} & \textbf{74.58} & \textbf{83.94} \\
16 & 91.45 & \underline{89.51} & \textbf{88.29} & \textbf{83.06} & 75.47 & 69.56 & 82.89 \\
32 & \textbf{92.77} & 88.12 & 86.27 & \underline{82.68} & 78.58 & \underline{73.92} & \underline{83.72} \\
64 & 89.83 & 87.17 & 85.62 & 81.67 & \underline{79.56} & 70.93 & 82.46 \\
128 & \underline{92.23} & \textbf{89.83} & \underline{86.36} & 81.25 & 74.38 & 71.90 & 82.66 \\
\bottomrule
\end{tabular*}
\end{table}

\FloatBarrier

\subsection{Long-Context Retrieval Analysis}
\label{sec:needle-in-a-haystack}

\begin{figure}[t!]
 \centering
 \newsavebox{\niahFigurePanel}
 \sbox{\niahFigurePanel}{%
  \begin{minipage}[t]{0.57\linewidth}
   \vspace{0pt}
   \setlength{\parskip}{0pt}
   \centering
   \includegraphics[width=\linewidth,trim=0 0 0 1.4bp,clip]{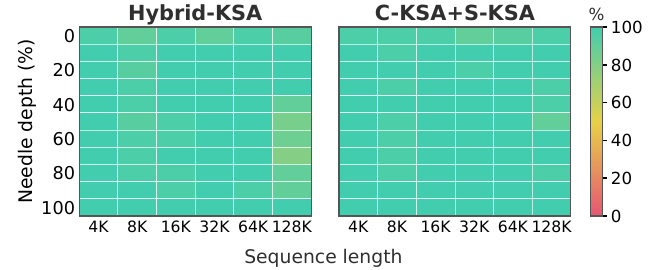}\par
   \captionsetup{type=figure,font=footnotesize,strut=off,skip=4pt}
   \caption{Single-needle retrieval across context lengths and needle depths.
   \textbf{C-KSA+S-KSA} maintains high accuracy across the grid, averaging \textbf{97.88\%} over all depths at 128K.}
   \label{fig:needle-in-a-haystack}
  \end{minipage}%
 }
 \begin{minipage}[t][\dimexpr\ht\niahFigurePanel+\dp\niahFigurePanel\relax][s]{0.40\linewidth}
  \vspace{0pt}
  \setlength{\parskip}{0pt}
  \captionsetup{type=table,font=footnotesize,skip=4pt}
  \caption{RULER-128K subtask scores (\%).
  \textbf{C-KSA+S-KSA} exceeds Hybrid KSA on Single, Multiquery, and SQuAD.}
  \label{tab:niah_subtasks}
  \vfill
  \centering
\begingroup
\footnotesize
\setlength{\tabcolsep}{2.5pt}
\renewcommand{\arraystretch}{1.05}
\begin{tabular}{@{}lcc@{}}
\toprule
Subtask & Hybrid KSA & C-KSA+S-KSA \\
\midrule
NIAH-Single & 93.33 & \textbf{98.33} \\
NIAH-Multikey & \textbf{85.00} & 83.33 \\
NIAH-Multivalue & \textbf{91.25} & 88.75 \\
NIAH-Multiquery & 91.25 & \textbf{96.25} \\
VT & \textbf{25.00} & 23.00 \\
FWE & \textbf{45.00} & 26.67 \\
SQuAD & 30.00 & \textbf{35.00} \\
\bottomrule
\end{tabular}
\endgroup
\par
  \vspace{0pt}
 \end{minipage}\hfill
 \usebox{\niahFigurePanel}
\end{figure}

We evaluate whether C-KSA+S-KSA retains the parent model's ability to retrieve information from different positions in a long context.
In the Needle-in-a-Haystack test~\citep{martin2023NIAH}, a target item is inserted into the context and retrieved through a targeted question.
Figure~\ref{fig:needle-in-a-haystack} compares the models on identical inputs across six context lengths from 4K to 128K and eleven insertion depths from 0\% to 100\%.
Table~\ref{tab:niah_subtasks} complements the grid with a paired comparison of RULER subtasks at 128K.

\FloatBarrier

\subsection{Inference Efficiency}
\label{sec:inference-efficiency}

C-KSA reduces inference costs by compressing remote Summary KV, while S-KSA replaces dense original-text attention with block-sparse access.
Figure~\ref{fig:quality-efficiency} summarizes the resulting compute and cache costs for C-KSA+S-KSA with $k_m=128$ and $b=4$.
At 128K, its estimated matrix-FLOP count per text token is lower by a factor of $6.38$ relative to Full and $2.76$ relative to Hybrid KSA.
Its BF16 KV and index-key cache is smaller by factors of $3.86$ and $1.35$, respectively.

\begin{figure}[htbp]
  \centering
  \includegraphics[width=.85\linewidth,trim=0 4bp 0 4bp,clip]{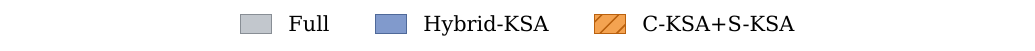}
  \par\vspace{-5pt}
  \includegraphics[width=.322\linewidth]{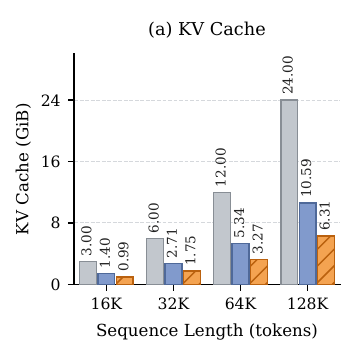}\hfill
  \includegraphics[width=.322\linewidth]{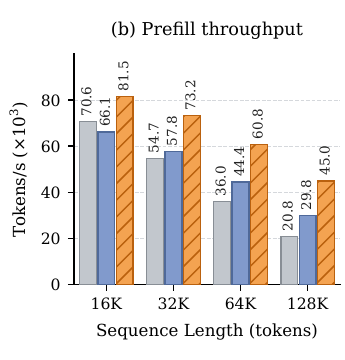}\hfill
  \includegraphics[width=.322\linewidth]{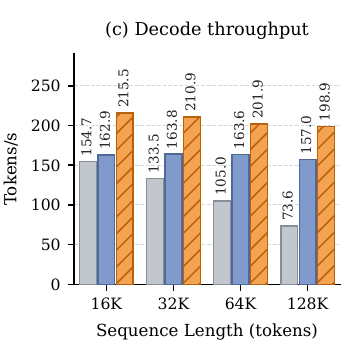}
  \caption{Inference cache size and throughput of \fourbrevision{1.9B} Full, Hybrid KSA, and C-KSA+S-KSA ($k_m=128$, $b=4$).}
  \label{fig:inference-efficiency}
\end{figure}

Figure~\ref{fig:inference-efficiency} compares the resident cache size, prefill throughput, and decode throughput of Full, Hybrid KSA, and C-KSA+S-KSA from 16K to 128K using \fourbrevision{the 1.9B models with} BF16 and batch size one on a single premier GPU.
C-KSA+S-KSA uses $k_m=128$ and $b=4$.

\textit{Cache.}
At 128K, C-KSA+S-KSA's resident cache occupies \textbf{6.31 GiB}, compared with 10.59 GiB for Hybrid KSA and 24.00 GiB for Full, reducing cache size by \textbf{40.4\%} and \textbf{73.7\%}, respectively.
Measurements include indexer, auxiliary state, KV storage, and resident workspaces.

\textit{Prefill.}
At 128K, C-KSA+S-KSA reaches \textbf{44,971 tokens/s}, compared with 29,837 for Hybrid KSA and 20,830 for Full, corresponding to \textbf{2.16$\times$} of Full attention.
Its gain over Hybrid KSA increases from $1.23\times$ at 16K to $1.51\times$ at 128K.

\textit{Decode.}
C-KSA+S-KSA achieves \textbf{215.50 tokens/s} at 16K and \textbf{198.85 tokens/s} at 128K, compared with 162.87 and 157.02 for Hybrid KSA, respectively.
Across the four sequence lengths, this gives a \textbf{1.23--1.32$\times$} speedup.

\FloatBarrier

\Needspace{7\baselineskip}
\subsection{Analysis and Ablation Studies}
\label{sec:ablation-studies}

\begingroup
\setlength{\emergencystretch}{2em}
\paragraph{Direct versus hierarchical compression.}
An alternative way to obtain $1024\times$ compression of distant-history KV is to increase the Summary chunk size directly to $k_s=1024$.
Table~\ref{tab:direct-hierarchical-compression} compares this approach with C-KSA+Full, using the original $k_s=8$ Hybrid KSA as a reference.
All three configurations retain Full-attention layers.
The large-chunk model is trained from scratch, while C-KSA builds on small-chunk Hybrid KSA.
\par\endgroup

\begin{table}[!htbp]
  \centering\small
  \setlength{\tabcolsep}{4pt}
  \caption{RULER performance of direct and hierarchical compression with Full attention retained.}
  \label{tab:direct-hierarchical-compression}
  \begin{tabular*}{\linewidth}{@{\extracolsep{\fill}}lccccccc@{}}
    \toprule
    Configuration & 4K & 8K & 16K & 32K & 64K & 128K & Mean \\
    \midrule
    Hybrid KSA, $k_s=8$ & \underline{80.65} & \textbf{73.35} & \underline{74.07} & \textbf{72.30} & \textbf{69.95} & 65.35 & \underline{72.61} \\
    Hybrid KSA, $k_s=1024$ & 75.65 & 72.14 & 70.44 & 70.81 & 69.11 & \underline{65.62} & 70.63 \\
    C-KSA+Full, $k_s=8$, $k_m=128$ & \textbf{81.52} & \underline{72.24} & \textbf{74.44} & \underline{71.93} & \underline{69.61} & \textbf{66.31} & \textbf{72.68} \\
    \bottomrule
  \end{tabular*}
\end{table}

Direct compression achieves a mean RULER score of 70.63, compared with 72.61 for the small-chunk reference and 72.68 for C-KSA+Full.
Hierarchical compression thus matches the reference performance and exceeds direct compression by 2.05 percentage points.
This finding is consistent with the 4B-model ablation in the KSA report~\citep[Table~7]{rw_ksa}, where increasing $k_s$ from 8 to 64 reduces the mean RULER score over 4K--64K from 82.97 to 77.79.
This advantage may stem from the \textbf{additional computation devoted to fine-grained Summary processing in C-KSA}, which may help preserve retrieval-relevant information at the same remote-memory compression ratio.
Actually, the additional computation as in two aspects: (1) the C-KSA takes 8 as the summary chunk, which will add more tokens at input sequence level; (2) the compressed KV memory is further computed by an additional module, which can be seen as a two-step summarization.

\begin{table}[t]
\centering\footnotesize
\setlength{\tabcolsep}{2pt}
\caption{Ablations of compression and sparse attention. Dashes indicate unreported scores since the sequence length is not enough to activate sparse mechanism.}
\label{tab:architectural-ablations}
\resizebox{\linewidth}{!}{%
\begin{tabular}{@{}l*{16}{c}@{}}
\toprule
\multirow{2}{*}{Configuration} & \multicolumn{7}{c}{Long-context retrieval (RULER)} & \multicolumn{4}{c}{Knowledge} & \multicolumn{3}{c}{Mathematics} & \multicolumn{2}{c}{Code} \\
\cmidrule(lr){2-8}\cmidrule(lr){9-12}\cmidrule(lr){13-15}\cmidrule(lr){16-17}
 & 4K & 8K & 16K & 32K & 64K & 128K & Mean & MMLU & CMMLU & C-Eval & \shortstack{MMLU\\Pro} & GSM8K & CMATH & MATH & MBPP & \shortstack{Human\\Eval} \\
\midrule
\rowcolor{gray!10}
\multicolumn{17}{l}{\textit{(a) C-KSA + Full: compression granularity}} \\
$k_m=4$ & 79.54 & 72.37 & 74.22 & 71.37 & 68.21 & 65.78 & 71.92 & 46.77 & 46.27 & \underline{43.76} & \underline{23.06} & \textbf{57.39} & \underline{64.00} & \underline{36.80} & 36.40 & 28.05 \\
$k_m=8$ & 79.42 & 72.56 & 74.35 & 71.69 & 68.95 & 66.15 & 72.19 & 46.78 & \underline{46.41} & \underline{43.76} & \textbf{23.12} & \underline{57.16} & \textbf{64.50} & 36.46 & \underline{36.60} & \underline{29.27} \\
$k_m=16$ & 79.95 & \textbf{73.31} & \textbf{74.92} & \textbf{72.31} & 68.46 & \textbf{67.46} & \textbf{72.74} & 46.95 & 46.30 & 43.39 & 22.61 & 56.71 & 63.17 & 36.56 & \textbf{37.40} & \textbf{31.10} \\
$k_m=32$ & 80.69 & \underline{72.73} & 74.04 & 71.86 & \underline{69.15} & 66.90 & 72.56 & \underline{47.04} & 46.30 & 43.46 & 22.83 & 56.33 & 62.67 & \textbf{37.06} & 36.20 & 28.66 \\
$k_m=64$ & \underline{80.94} & 72.00 & 73.68 & \underline{72.30} & 68.41 & \underline{67.04} & 72.40 & \textbf{47.12} & 46.19 & 43.46 & 22.92 & 56.25 & 63.50 & 36.70 & 36.00 & 27.44 \\
$k_m=128$ (selected) & \textbf{81.52} & 72.24 & \underline{74.44} & 71.93 & \textbf{69.61} & 66.31 & \underline{72.68} & 46.97 & \textbf{46.49} & \textbf{43.83} & 22.71 & 55.95 & \textbf{64.50} & 36.60 & 35.60 & 28.05 \\
\midrule
\rowcolor{gray!10}
\multicolumn{17}{l}{\textit{(b) C-KSA, $k_m=128$: sparse attention and teacher pooling}} \\
DSA (token routing) & \underline{81.38} & 72.22 & 75.12 & \underline{71.22} & \textbf{69.37} & \textbf{65.37} & \underline{72.45} & \textemdash & \textemdash & \textemdash & \textemdash & \textemdash & \textemdash & \textemdash & \textemdash & \textemdash \\
QSA, $b=4$, max & \textbf{81.48} & 72.19 & 75.15 & 70.26 & 67.57 & 61.39 & 71.34 & \textemdash & \textemdash & \textemdash & \textemdash & \textemdash & \textemdash & \textemdash & \textemdash & \textemdash \\
S-KSA, $b=2$, max & 81.08 & \textbf{72.67} & \underline{75.30} & \textbf{71.58} & \underline{69.21} & \underline{65.23} & \textbf{72.51} & \textemdash & \textemdash & \textemdash & \textemdash & \textemdash & \textemdash & \textemdash & \textemdash & \textemdash \\
S-KSA, $b=4$, max (selected) & \textbf{81.48} & 72.42 & \textbf{75.67} & 70.78 & 68.99 & 65.01 & 72.39 & \textemdash & \textemdash & \textemdash & \textemdash & \textemdash & \textemdash & \textemdash & \textemdash & \textemdash \\
S-KSA, $b=8$, max & \underline{81.38} & \underline{72.47} & 74.15 & 68.29 & 64.83 & 57.62 & 69.79 & \textemdash & \textemdash & \textemdash & \textemdash & \textemdash & \textemdash & \textemdash & \textemdash & \textemdash \\
S-KSA, $b=4$, sum & \textbf{81.48} & 72.15 & 74.81 & 70.28 & 68.51 & 61.68 & 71.49 & \textemdash & \textemdash & \textemdash & \textemdash & \textemdash & \textemdash & \textemdash & \textemdash & \textemdash \\
\midrule
\rowcolor{gray!10}
\multicolumn{17}{l}{\textit{(c) Backbone for sparse adaptation: S-KSA, $b=4$, max teacher}} \\
KSA + S-KSA & \underline{80.71} & \textbf{73.82} & \underline{74.76} & \textbf{72.12} & \underline{68.57} & \underline{61.06} & \underline{71.84} & \underline{46.60} & \textbf{46.99} & \textbf{43.83} & \underline{22.62} & \textbf{58.83} & \underline{61.67} & \textbf{37.40} & \textbf{36.80} & \textbf{31.71} \\
C-KSA + S-KSA, $k_m=128$ & \textbf{81.48} & \underline{72.42} & \textbf{75.67} & \underline{70.78} & \textbf{68.99} & \textbf{65.01} & \textbf{72.39} & \textbf{47.09} & \underline{46.39} & \underline{42.94} & \textbf{22.83} & \underline{56.33} & \textbf{63.50} & \underline{36.72} & \underline{35.20} & \underline{29.88} \\
\bottomrule
\end{tabular}
}
\end{table}

\paragraph{Compression granularity with Full attention.}
Table~\ref{tab:architectural-ablations}(a) compares $k_m\in\{4,8,16,32,64,128\}$ in C-KSA+Full.
The six-length RULER mean ranges from 71.92 to 72.74, with no systematic decline as compression increases.
The Full layers may help preserve retrieval quality by providing access to original-text details lost during compression.

At 128K, $k_m=128$ leaves about 128 Meta entries per C-KSA layer, so larger $k_m$ would save little further cache; we therefore select $k_m=128$ and leave higher factors to million-token contexts.

\paragraph{Block size and routing methods.}
Table~\ref{tab:architectural-ablations}(b) compares sparse-attention schemes on the same $k_m=128$ C-KSA+Full backbone, including S-KSA with $b\in\{2,4,8\}$, token-level DSA~\citep{rw_deepseekv32}, and Qwen Sparse Attention (QSA)~\citep{rw_qsa}.
QSA provides a simple alternative to Summary-guided aggregation: it averages the projected token-level index keys within each block.
We follow Qwen's QSA indexer design, using $b=4$ and eight indexer heads to match our comparison settings.
All methods are compared under the same experimental setup.
We report only RULER in this panel, since most general-capability sequences fit within the combined remote and local attention budget and offer limited discrimination of routing quality.

S-KSA with $b=4$ achieves a mean RULER score comparable to those of $b=2$ and token-level DSA, with only half and one quarter as many routing candidates, respectively.
It also outperforms QSA at the same block size.
Increasing to $b=8$ reduces the candidate count further but degrades retrieval quality, particularly at 128K.
We therefore select $b=4$ as a balance between retrieval quality and routing cost.

\paragraph{Max versus sum teacher pooling.}
At $k_m=128$ and $b=4$, we compare the max-pooled teacher in Equation~\ref{eq:indexer-loss}, which follows QSA~\citep{rw_qsa}, with a sum-pooled teacher that uses $\sum_{i\in\mathcal B_u}a_{t,i}$ as each block's unnormalized weight.
Both runs use the same training setup, differing only in teacher pooling.
Max pooling improves the mean RULER score from 71.49\% to 72.39\% and the 128K score from 61.68\% to 65.01\% (Table~\ref{tab:architectural-ablations}(b)).
Retrieval likely depends on a few salient tokens, whose signal sum pooling dilutes with diffuse attention.
These results confirm QSA's choice on the KSA backbone.

\paragraph{Failure mode.}
The largest 128K subtask decrease is frequent-word extraction, from 45.00\% to 26.67\% (Table~\ref{tab:niah_subtasks}), an aggregation task.

\paragraph{Backbone for sparse adaptation.}
Table~\ref{tab:architectural-ablations}(c) compares S-KSA adapted on Hybrid KSA and C-KSA+Full with $k_m=128$.
Both use $b=4$, a max-pooled teacher, and 18.5B tokens of module-only sparse adaptation with the same remote budget, local window, and sink settings.
C-KSA+S-KSA achieves a mean RULER score of 72.39 and a 128K score of 65.01, compared with 71.84 and 61.06 for KSA+S-KSA.
Relative to their respective dense references, the 128K reduction is 1.30 points for C-KSA versus 4.29 for Hybrid KSA, indicating better retention of the backbone's retrieval performance in this comparison.

\begin{figure}[t]
 \centering
 \includegraphics[width=\linewidth]{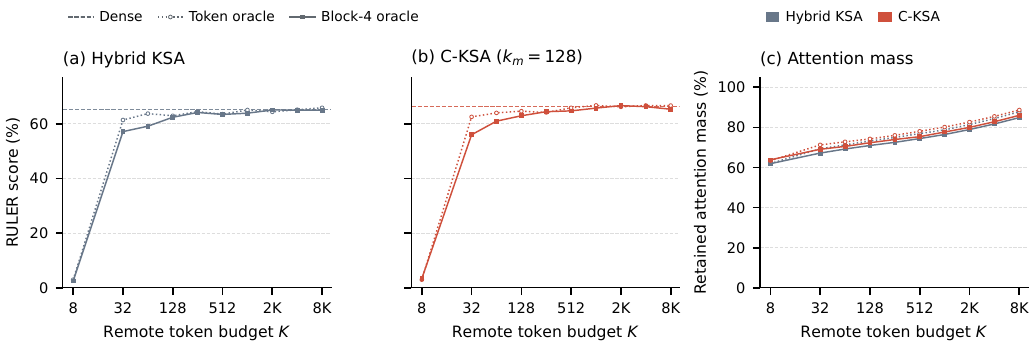}
 \caption{Oracle selection at 128K with frozen Hybrid KSA (blue) and C-KSA+Full ($k_m=128$, red).
 (a,b) RULER scores.
 (c) Decode attention mass retained by token and block selection ($b=4$), including local context and sinks before sparse renormalization.}
 \label{fig:routing-budget}
\end{figure}

Figure~\ref{fig:routing-budget}(c) provides a complementary observation: at the same token budget, oracle block selection retains slightly more decode attention mass on C-KSA than on Hybrid KSA, including local context and sinks.
This is consistent with the smaller retrieval loss after sparse adaptation and suggests that the compressed backbone remains well suited to sparse original-text access.
Together, these results support the compatibility of hierarchical compression and sparse attention.

\paragraph{Sparse selection using Full attention scores.}
To test whether sparse attention requires backbone adaptation, we freeze all model parameters and select remote tokens or blocks directly using Full attention probabilities averaged across heads.
Token selection ranks individual probabilities; block selection ranks the maximum probability within each block of $b=4$ tokens, matching the Indexer's teacher pooling rule.
We call this \emph{oracle} selection and compare it with the same backbone's unchanged Full layers.
The remote budget $K$ includes eight sink tokens; a 1024-token local window is retained separately.

\begin{table}[t]
\centering\small
\caption{Dense and oracle RULER results with frozen backbones. Oracle selection uses $K=4096$; Mean averages all six context lengths (4K--128K).}
\label{tab:oracle-quality}
\setlength{\tabcolsep}{4pt}
\begin{tabular}{lcccccc}
\toprule
\multirow{2}{*}{Backbone} & \multicolumn{2}{c}{Dense} & \multicolumn{2}{c}{Token oracle} & \multicolumn{2}{c}{Block-4 oracle} \\
\cmidrule(lr){2-3}\cmidrule(lr){4-5}\cmidrule(lr){6-7}
 & Mean & 128K & Mean & 128K & Mean & 128K \\
\midrule
Hybrid KSA & 72.61 & 65.35 & 73.12 & 65.13 & 73.05 & 65.00 \\
C-KSA+Full, $k_m=16$ & 72.74 & 67.46 & 73.24 & 68.94 & 73.17 & 69.41 \\
C-KSA+Full, $k_m=32$ & 72.56 & 66.90 & 72.69 & 66.93 & 72.81 & 67.22 \\
C-KSA+Full, $k_m=64$ & 72.40 & 67.04 & 72.87 & 66.56 & 72.63 & 66.89 \\
C-KSA+Full, $k_m=128$ & 72.68 & 66.31 & 72.84 & 66.67 & 72.66 & 66.24 \\
\bottomrule
\end{tabular}
\end{table}

Figure~\ref{fig:routing-budget} shows rapid recovery toward dense performance as the remote budget increases on both Hybrid KSA and C-KSA with $k_m=128$.
On C-KSA, block-level selection reaches 64.71 at $K=512$.
At $K=4096$, the selected remote tokens and the 1024-token local window together cover only about 3.9\% of a 128K context, yet retain 82.83\% of decode attention mass.
The corresponding RULER score is 66.24, compared with 66.31 for dense attention.
This concentration of attention on a small subset of tokens helps explain why sparse attention can preserve retrieval performance when relevant tokens are selected.
At the same budget, Hybrid KSA retains 81.71\% of decode attention mass.
Local context and sinks alone already retain substantial attention mass but yield poor retrieval performance, highlighting the importance of selecting relevant remote content.

Table~\ref{tab:oracle-quality} extends the comparison to Hybrid KSA and four C-KSA+Full backbones at $K=4096$, reporting mean RULER performance across all six context lengths and the 128K score.
Both selection methods remain within 0.51 percentage points of the dense references in mean RULER score across all five backbones.
Accurate selection therefore recovers near-dense RULER performance without backbone updates in these settings, including on models already adapted to C-KSA.

Full-parameter sparse continuation also showed no clear RULER improvement under the training recipe and budget we tested.
Together with the recovery observed after module-only training in Section~\ref{sec:model-quality}, these results support focusing sparse adaptation on the aggregation module and Indexer rather than requiring backbone updates.

\begin{figure}[p]
 \centering
 \includegraphics[width=\linewidth]{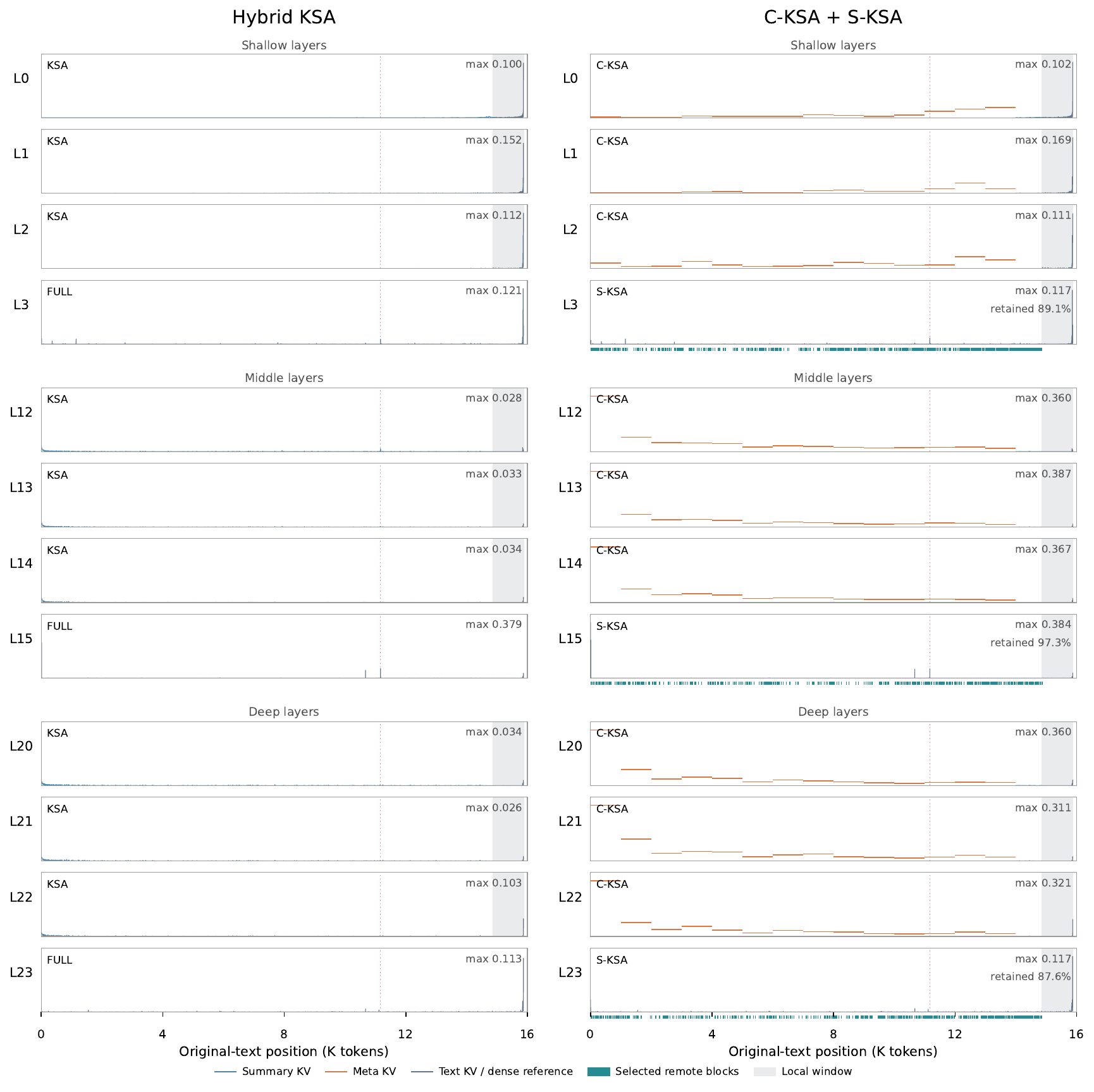}
 \caption{Per-layer attention and block selection on a shared 16K NIAH example for Hybrid KSA (left) and C-KSA+S-KSA (right).
 Teal strips mark selected remote blocks, dashed lines mark the answer, and gray shading denotes the local window.}
 \label{fig:attention-patterns}
\end{figure}

\paragraph{Attention patterns and block selection.}
Figure~\ref{fig:attention-patterns} compares layer-wise attention in Hybrid KSA and our main C-KSA+S-KSA model on a shared 16K NIAH example, using $k_m=128$, $b=4$, and $K=4096$ for the latter.
Both models correctly retrieve the answer from outside the local window.
We visualize attention when predicting the first answer digit.
For C-KSA, the plotted attention probabilities include the $\log k_m$ correction to Meta logits before softmax.
For S-KSA, we plot dense attention probabilities and mark the text blocks selected by the Indexer.

For the visualized query, Meta entries receive 7.7--13.7\% of C-KSA's attention mass in the displayed shallow layers, compared with 73.1--87.8\% in the middle and deep layers, showing substantial use of compressed remote memory.
In all three displayed S-KSA layers, the selected blocks cover the answer span.
Together with local context, these blocks retain 87.6--97.3\% of the dense attention mass while accessing only 31.5\% of the original-text positions.

\FloatBarrier
\noindent
\begin{minipage}[t]{0.48\linewidth}
\vspace{0pt}
\paragraph{Continued pretraining of Hybrid KSA.}
Because our adapted models receive additional training while the reported baselines do not, we examine whether continued pretraining of the unchanged Hybrid KSA backbone would provide a stronger quality reference.
Table~\ref{tab:hybrid-ksa-continuation} reports a full-parameter continuation experiment from the same 128K checkpoint, using a global batch size of 256, a peak learning rate of $4\times10^{-4}$, and a warmup--stable--decay schedule.
Across five evaluated checkpoints covering 5.7--51.3B additional tokens, the six-length RULER mean remains below the original checkpoint's 72.60.
The 128K score improves at 28.5B tokens but falls back at later checkpoints.
Under the tested recipe and data budgets, continued pretraining does not yield a consistent improvement in Hybrid KSA's RULER performance.
This supports using the original checkpoint as the quality reference for assessing whether architectural adaptation can preserve its capabilities at lower inference cost.
\end{minipage}\hfill
\begin{minipage}[t]{0.48\linewidth}
\vspace{0pt}
\begingroup
  \captionsetup{type=table}
  \centering
  \caption{RULER scores after continued pretraining of Hybrid KSA.}
  \label{tab:hybrid-ksa-continuation}
  \small
  \renewcommand{\arraystretch}{1.15}
  \begin{tabular*}{\linewidth}{@{\extracolsep{\fill}}lrr@{}}
    \toprule
    Added tokens (B) & Mean & 128K \\
    \midrule
    0 (original) & \textbf{72.60} & \underline{65.27} \\
    \midrule
    5.7 & 70.84 & 63.26 \\
    17.1 & 70.77 & 63.72 \\
    28.5 & 71.86 & \textbf{66.76} \\
    39.9 & 69.74 & 63.17 \\
    51.3 & \underline{71.95} & 63.62 \\
    \bottomrule
  \end{tabular*}
  \par
\endgroup

\end{minipage}\par

\section{Related Work}
\label{sec:related-work}

Multi-head attention (MHA) is a fundamental component of Transformer language models, with separate query, key, and value projections for each head~\citep{rw_attention}.
Full attention provides direct access to preceding tokens, but its computation during training and prefill grows quadratically with sequence length, while its KV cache grows linearly during autoregressive decoding.
FlashAttention reduces memory traffic for exact attention~\citep{rw_flashattention}, while PagedAttention improves KV-cache allocation and sharing~\citep{rw_pagedattention}.
These system-level optimizations complement research on memory compression, sparse attention, and hybrid architectures.

\subsection{Memory Compression}
\label{sec:related-memory-compression}

One line of memory compression reduces storage at each token position.
Multi-query attention (MQA) shares a single KV head across all query heads~\citep{rw_mqa}, while grouped-query attention (GQA) shares KV heads within groups of query heads~\citep{rw_gqa}.
Multi-head latent attention (MLA) stores compact per-token latent states~\citep{rw_deepseekv2}, and quantization methods such as KIVI reduce the precision of cached entries~\citep{rw_kivi}.
These approaches reduce storage per position without removing historical positions.
Other methods retain only a subset of token-level KV entries: H2O~\citep{rw_h2o} and SnapKV~\citep{rw_snapkv} use attention statistics to identify important positions, while StreamingLLM~\citep{rw_streamingllm} retains initial attention sinks together with a recent window.

Another direction compresses historical context into fixed-size recurrent states rather than retaining token-level KV entries.
Linear Transformers~\citep{rw_linear_attention} achieve this through kernel feature maps and recurrent accumulation of key--value statistics.
Subsequent designs introduce decay in RetNet~\citep{rw_retnet}, data-dependent gating in GLA~\citep{rw_gla}, and gated delta-rule updates in Gated DeltaNet~\citep{rw_gated_deltanet}.
Kimi Delta Attention (KDA)~\citep{rw_kimi_linear} further extends Gated DeltaNet with finer-grained gating for recurrent memory updates.
Related approaches include RWKV~\citep{peng2024eagle}, with matrix-valued recurrent states, and selective state-space models such as Mamba~\citep{rw_mamba} and Mamba-2~\citep{rw_mamba2}.
In their recurrent forms, these models maintain a state size independent of context length, but access historical information through aggregated states rather than separately retained memory entries.

Sequence-level aggregation represents multiple context positions with fewer memory entries that remain accessible to attention.
Compressive Transformer~\citep{rw_compressive_transformer} compresses older hidden-state memories, while Dynamic Memory Compression~\citep{rw_dmc} learns adaptive KV-cache compression.
Learned summary and memory representations are explored in Gist~\citep{rw_gist}, ICAE~\citep{rw_icae}, and UniGist~\citep{rw_unigist}.
AutoCompressors~\citep{chevalier2023adapting} uses summary vectors as soft prompts, whereas Activation Beacon~\citep{rw_activation_beacon} retains compressed KV activations at each layer.
GMSA~\citep{rw_gmsa} combines group merging and layer semantic alignment in an encoder--decoder context-compression framework.
Cartridges~\citep{rw_cartridges} instead optimizes corpus-specific KV representations offline for reuse across subsequent queries.

KSA~\citep{rw_ksa} interleaves Summary tokens with fixed-size text chunks.
In KSA layers, text queries directly access nearby text and retrieve distant context through Summary KV, while the Summary hidden states continue to be updated across layers.
A systematic study of gist-based compression identifies task-dependent limitations, particularly in synthetic recall~\citep{rw_gist_study}.
These findings underscore the importance of evaluating retained capabilities rather than compression ratio alone.

Other designs organize compressed memory at multiple resolutions.
MELODI~\citep{rw_melodi} combines recurrent short-term memory with further-compressed long-term memory constructed at an intermediate layer.
AdmTree~\citep{li2025admtree} organizes gist representations into adaptive semantic trees for hierarchical context compression.
H-SSA~\citep{rw_ssa} inserts meta-gist tokens to summarize groups of gist tokens and support coarse-to-fine retrieval.
DeepSeek-V4~\citep{rw_deepseekv4} applies learned dimension-wise aggregation to text-token KV, using different compression granularities in Compressed Sparse Attention (CSA) and Heavily Compressed Attention (HCA).
C-KSA instead applies dimension-wise aggregation to Summary KV, forming Meta entries used only within the layer that constructs them, rather than introducing additional tokens into the hidden-state sequence.
This further reduces distant-history storage without coarsening the Summary sequence passed between layers, and permits each layer to choose its own KV compression ratio.

\subsection{Sparse Attention}
\label{sec:related-sparse-attention}

Sparse attention reduces computation by restricting the positions read by each query.
Sparse Transformer~\citep{rw_sparse_transformer} uses structured sparse factorizations, Longformer~\citep{rw_longformer} combines local windows with global attention, and BigBird~\citep{rw_bigbird} additionally incorporates random connections.
Sliding-window attention, used in Mistral~\citep{rw_mistral}, restricts each layer's direct access to nearby tokens.
Another line exploits attention sparsity in pretrained models without additional training.
QUEST~\citep{rw_quest} estimates the relevance of KV-cache pages to each query and reads only the selected pages.
MInference~\citep{rw_minference} combines head-specific pattern templates with input-dependent sparse indices to accelerate prefill, while SpargeAttention~\citep{rw_spargeattn} and XAttention~\citep{rw_xattention} estimate attention importance to skip low-contribution computations.
These methods sparsify attention through prescribed patterns or inference-time selection, without training an auxiliary routing module.

Content-dependent sparse architectures integrate routing into the attention mechanism.
SeerAttention~\citep{rw_seerattention} learns block-level gates by self-distilling dense attention, while DeepSeek Sparse Attention (DSA)~\citep{rw_deepseekv32} uses a Lightning Indexer to select individual tokens.
At block granularity, MoBA~\citep{rw_moba} scores mean-pooled token keys, whereas Qwen Sparse Attention (QSA)~\citep{rw_qsa} mean-pools auxiliary index keys within micro-blocks; both attend to the selected blocks' original-token KV.
Native Sparse Attention (NSA)~\citep{rw_nsa} uses learned block compression to guide fine-grained selection and combines compressed, selected, and local attention branches.
In contrast, DeepSeek-V4's CSA~\citep{rw_deepseekv4} selects compressed KV entries for its remote attention.
S-KSA also reduces routing to block granularity, but reuses existing KSA Summaries as aggregation queries to construct routing representations and retains original-token KV for the main attention.

Learned memory tokens can also provide an interface for block retrieval.
Landmark Attention~\citep{rw_landmark} represents each context block with a landmark token and learns to retrieve blocks through attention to these landmarks.
Simplified Sparse Attention (SSA)~\citep{rw_ssa} scores gist keys and unfolds selected chunks to recover their original tokens.
HiLS-Attention~\citep{rw_hils} uses landmark queries to pool token keys, then combines within-chunk attention outputs using entropy-corrected chunk weights.
This allows the language-modeling loss to train routing; SAS~\citep{rw_sas} similarly learns context ranking through differentiable gates inside attention normalization.
In S-KSA, Summaries guide auxiliary aggregation rather than serving directly as routing keys.
We train the aggregation modules and Indexer by attention distillation and use the resulting scores only for block selection.
The main attention applies a single softmax over selected original tokens and local context, without using routing scores as additional attention weights.

\subsection{Hybrid Architectures}
\label{sec:related-hybrid}

Hybrid architectures seek to retain the fine-grained retrieval of attention while reducing the cost of processing long contexts.
A common strategy interleaves several efficient layers with a smaller number of full-attention layers.
Hybrid H3~\citep{rw_h3}, Mamba-2-Hybrid~\citep{rw_mamba_hybrid}, and Jamba~\citep{rw_jamba} combine state-space layers with attention, while Kimi Linear~\citep{rw_kimi_linear} combines KDA with MLA.
Local attention provides another complementary component: RecurrentGemma~\citep{botev2024recurrentgemma} and Samba~\citep{rw_samba} pair it with recurrent or state-space layers, whereas Gemma~3~\citep{rw_gemma3} and MiMo-V2-Flash~\citep{rw_mimo2} interleave it with global attention.

Beyond layer-wise interleaving, other designs exploit sharing or finer-grained mixing.
Zamba~\citep{rw_zamba} reuses a shared attention module at multiple depths of a Mamba backbone.
YOCO~\citep{rw_yoco} stacks a cross-decoder on an efficient self-decoder and shares a global KV cache across cross-decoder layers.
HySparse~\citep{rw_hysparse} reuses token selections and global KV from full-attention layers in subsequent sparse layers, while DeepSeek-V4.1~\citep{rw_deepseekv41} introduces cross-layer KV and index reuse within compressed sparse attention.
At a finer granularity, Hymba~\citep{rw_hymba} runs attention and state-space heads in parallel within the same layer.

Hybrid KSA~\citep{rw_ksa} combines Summary-based KSA layers with full-attention layers in a $3{:}1$ layout.
KSA V2 retains this layer-wise structure while improving both components: C-KSA further compresses the Summary KV used for distant-history access, and S-KSA replaces dense attention over original text with Summary-guided block-sparse attention.

\section{Conclusion and Future Work}
\label{sec:conclusion}

We showed that the Summaries of a summary-token hybrid can serve both as compressible memory and as a routing index.
C-KSA compresses Summary KV within each layer without coarsening the Summaries passed between layers, and S-KSA replaces full attention with Summary-guided block-sparse attention over original text.
Retrofitted into a pretrained \fourbrevision{1.9B} Hybrid KSA model, KSA V2 retains its retrieval performance and largely preserves general capabilities, while reducing cache size and improving inference throughput.
\fourbrevision{Additional experiments on a 4B GQA backbone retain near-parent RULER performance, supporting transfer to a larger model with shared KV heads.}
The remaining cost lies mainly in the original-text KV stored by the sparse layers, a natural target for future work.

\fourbrevision{Our evaluation covers 1.9B MHA and 4B GQA backbones with contexts up to 128K. General-capability evaluations and efficiency measurements are limited to 1.9B; the 4B validation covers RULER retrieval.}
Future work will explore longer contexts and larger models, including whether additional compression levels can provide further savings while preserving useful information.
We will also explore how to allocate compression factors across layers.
Evaluating long-document understanding and multi-step reasoning will also help establish how well the observed retrieval performance transfers to broader long-context tasks.


\newpage
\section{Author List and Acknowledgement}
\noindent
\textbf{Core Contributors}\quad
Chenglong Chu,
Guohong Mu,
Guowang Zhang,
Hao Peng,
Hui Wang,
Jian Liang,
Jiangxia Cao\textsuperscript{$\dagger$},
Jiao Ou,
Jiaxin Deng,
Lu Ren\textsuperscript{*},
Muhao Wei,
Ruiming Tang\textsuperscript{*},
Tianle Xu,
Wenwu Ou,
Xingmei Wang,
Zhixin Ling,
Zhiyuan Liang\textsuperscript{*},
Ziqi Wang\textsuperscript{*}

\vspace{0.5em}
\noindent
\textbf{Contributors}\quad
Chaoyi Ma,
Haixiao Gao,
Kuo Cai,
Renpu Liu,
Ruochen Yang,
Wenlong Xu,
Xinqi Jin,
Yan Sun,
Yifan Wang,
Yingzhi He,
Yiyu Wang,
Yusen Huo,
Zhaojie Liu,
Zihan Zhou

\vspace{0.5em}
\noindent
{\small All the authors listed alphabetically by first name.\quad \quad \textsuperscript{*}individuals who have departed from our team.\\
\textsuperscript{$\dagger$}Project Leader: caojiangxia@kuaishou.com}

\clearpage
\printbibliography[title={References}]

@inproceedings{rw_attention,
  title = {{Attention is All you Need}},
  author = {Vaswani, Ashish and Shazeer, Noam and Parmar, Niki and Uszkoreit, Jakob and Jones, Llion and Gomez, Aidan N. and Kaiser, Łukasz and Polosukhin, Illia},
  year = {2017},
  booktitle = {Advances in Neural Information Processing Systems},
  volume = {30},
  url = {https://proceedings.neurips.cc/paper_files/paper/2017/hash/3f5ee243547dee91fbd053c1c4a845aa-Abstract.html}
}

@article{rw_rope,
  title = {{RoFormer: Enhanced transformer with Rotary Position Embedding}},
  author = {Jianlin Su and Murtadha Ahmed and Yu Lu and Shengfeng Pan and Wen Bo and Yunfeng Liu},
  year = {2024},
  journal = {Neurocomputing},
  doi = {10.1016/j.neucom.2023.127063},
  volume = {568},
  pages = {127063},
  url = {https://www.sciencedirect.com/science/article/pii/S0925231223011864}
}

@inproceedings{rw_ruler,
  title = {{RULER: What's the Real Context Size of Your Long-Context Language Models?}},
  author = {Cheng-Ping Hsieh and Simeng Sun and Samuel Kriman and Shantanu Acharya and Dima Rekesh and Fei Jia and Yang Zhang and Boris Ginsburg},
  year = {2024},
  booktitle = {Conference on Language Modeling},
  url = {https://openreview.net/forum?id=kIoBbc76Sy}
}

@article{rw_mqa,
  title = {{Fast Transformer Decoding: One Write-Head is All You Need}},
  author = {Shazeer, Noam},
  year = {2019},
  journal = {arXiv preprint arXiv:1911.02150},
  url = {https://arxiv.org/abs/1911.02150}
}

@inproceedings{rw_gqa,
  title = {{GQA: Training Generalized Multi-Query Transformer Models from Multi-Head Checkpoints}},
  author = {Joshua Ainslie and James Lee-Thorp and Michiel de Jong and Yury Zemlyanskiy and Federico Lebron and Sumit Sanghai},
  year = {2023},
  booktitle = {Proceedings of the 2023 Conference on Empirical Methods in Natural Language Processing},
  doi = {10.18653/v1/2023.emnlp-main.298},
  pages = {4895--4901},
  url = {https://aclanthology.org/2023.emnlp-main.298/}
}

@article{rw_deepseekv2,
  title = {{DeepSeek-V2: A Strong, Economical, and Efficient Mixture-of-Experts Language Model}},
  author = {{DeepSeek-AI} and Liu, Aixin and Feng, Bei and Wang, Bin and Wang, Bingxuan and Liu, Bo and Zhao, Chenggang and Dengr, Chengqi and Ruan, Chong and others},
  year = {2024},
  journal = {arXiv preprint arXiv:2405.04434},
  url = {https://arxiv.org/abs/2405.04434}
}

@inproceedings{rw_yoco,
  title = {{You Only Cache Once: Decoder-Decoder Architectures for Language Models}},
  author = {Sun, Yutao and Dong, Li and Zhu, Yi and Huang, Shaohan and Wang, Wenhui and Ma, Shuming and Zhang, Quanlu and Wang, Jianyong and Wei, Furu},
  year = {2024},
  booktitle = {Advances in Neural Information Processing Systems},
  volume = {37},
  doi = {10.52202/079017-0235},
  pages = {7339--7361},
  url = {https://proceedings.neurips.cc/paper_files/paper/2024/hash/0df38cd13520747e1e64e5b123a78ef8-Abstract-Conference.html}
}

@inproceedings{rw_compressive_transformer,
  title = {{Compressive Transformers for Long-Range Sequence Modelling}},
  author = {Rae, Jack W. and Potapenko, Anna and Jayakumar, Siddhant M. and Hillier, Chloe and Lillicrap, Timothy P.},
  year = {2020},
  booktitle = {International Conference on Learning Representations},
  url = {https://openreview.net/forum?id=SylKikSYDH}
}

@inproceedings{rw_gist,
  title = {{Learning to Compress Prompts with Gist Tokens}},
  author = {Mu, Jesse and Li, Xiang and Goodman, Noah},
  year = {2023},
  booktitle = {Advances in Neural Information Processing Systems},
  volume = {36},
  doi = {10.52202/075280-0848},
  pages = {19327--19352},
  url = {https://proceedings.neurips.cc/paper_files/paper/2023/hash/3d77c6dcc7f143aa2154e7f4d5e22d68-Abstract-Conference.html}
}

@inproceedings{rw_landmark,
  title = {{Random-Access Infinite Context Length for Transformers}},
  author = {Mohtashami, Amirkeivan and Jaggi, Martin},
  year = {2023},
  booktitle = {Advances in Neural Information Processing Systems},
  volume = {36},
  doi = {10.52202/075280-2378},
  pages = {54567--54585},
  url = {https://proceedings.neurips.cc/paper_files/paper/2023/hash/ab05dc8bf36a9f66edbff6992ec86f56-Abstract-Conference.html}
}

@inproceedings{rw_icae,
  title = {{In-context Autoencoder for Context Compression in a Large Language Model}},
  author = {Ge, Tao and Jing, Hu and Wang, Lei and Wang, Xun and Chen, Si-Qing and Wei, Furu},
  year = {2024},
  booktitle = {International Conference on Learning Representations},
  pages = {2591--2607},
  url = {https://proceedings.iclr.cc/paper_files/paper/2024/hash/0b276510ec2d3f6613a8b60c41ff0438-Abstract-Conference.html}
}

@inproceedings{rw_activation_beacon,
  title = {{Long Context Compression with Activation Beacon}},
  author = {Zhang, Peitian and Liu, Zheng and Xiao, Shitao and Shao, Ninglu and Ye, Qiwei and Dou, Zhicheng},
  year = {2025},
  booktitle = {International Conference on Learning Representations},
  pages = {101932--101948},
  url = {https://proceedings.iclr.cc/paper_files/paper/2025/hash/fc797c61eb18f7b3ce9a74af0ef0d876-Abstract-Conference.html}
}

@article{rw_ksa,
  title = {{Kwai Summary Attention Technical Report}},
  author = {Chu, Chenglong and Zhou, Guorui and Zhang, Guowang and Li, Han and Peng, Hao and Cheng, Hongtao and Wang, Hui and Liang, Jian and Cao, Jiangxia and others},
  year = {2026},
  journal = {arXiv preprint arXiv:2604.24432},
  url = {https://arxiv.org/abs/2604.24432}
}

@inproceedings{rw_dmc,
  title = {{Dynamic Memory Compression: Retrofitting LLMs for Accelerated Inference}},
  author = {Piotr Nawrot and Adrian Łańcucki and Marcin Chochowski and David Tarjan and Edoardo Ponti},
  year = {2024},
  booktitle = {International Conference on Machine Learning},
  publisher = {PMLR},
  pages = {37396--37412},
  url = {https://proceedings.mlr.press/v235/nawrot24a.html}
}

@inproceedings{rw_melodi,
  title = {{MELODI: Exploring Memory Compression for Long Contexts}},
  author = {Chen, Yinpeng and Hutchins, DeLesley and Jansen, Aren and Zhmoginov, Andrey and Racz, David and Andersen, Jesper},
  year = {2025},
  booktitle = {International Conference on Learning Representations},
  pages = {67589--67611},
  url = {https://proceedings.iclr.cc/paper_files/paper/2025/hash/a8e3197f627b9c88f86c0d3eb1ade5d7-Abstract-Conference.html}
}

@inproceedings{rw_unigist,
  title = {{UniGist: Towards General and Hardware-aligned Sequence-level Long Context Compression}},
  author = {Deng, Chenlong and Zhang, Zhisong and Mao, Kelong and Li, Shuaiyi and Fang, Tianqing and Zhang, Hongming and Mi, Haitao and Yu, Dong and Dou, Zhicheng},
  year = {2025},
  booktitle = {Advances in Neural Information Processing Systems},
  volume = {38},
  doi = {10.52202/085713-0665},
  pages = {19775--19797},
  url = {https://proceedings.neurips.cc/paper_files/paper/2025/hash/1c93b738747776f9d3fdd077a5a07114-Abstract-Conference.html}
}

@inproceedings{rw_gist_study,
  title = {{A Silver Bullet or a Compromise for Full Attention? A Comprehensive Study of Gist Token-based Context Compression}},
  author = {Chenlong Deng and Zhisong Zhang and Kelong Mao and Shuaiyi Li and Xinting Huang and Dong Yu and Zhicheng Dou},
  year = {2025},
  booktitle = {Proceedings of the 63rd Annual Meeting of the Association for Computational Linguistics (Volume 1: Long Papers)},
  doi = {10.18653/v1/2025.acl-long.241},
  pages = {4861--4879},
  url = {https://aclanthology.org/2025.acl-long.241/}
}

@inproceedings{rw_cartridges,
  title = {{Cartridges: Lightweight and general-purpose long context representations via self-study}},
  author = {Eyuboglu, Sabri and Ehrlich, Ryan and Arora, Simran and Guha, Neel and Zinsley, Dylan and Liu, Emily and Rudra, Atri and Zou, James Y and Mirhoseini, Azalia and Re, Christopher},
  year = {2026},
  booktitle = {International Conference on Learning Representations},
  pages = {42642--42687},
  url = {https://proceedings.iclr.cc/paper_files/paper/2026/hash/4681359a7b1e94571598ad1adda35e6e-Abstract-Conference.html}
}

@inproceedings{rw_gmsa,
  title = {{GMSA: Enhancing Context Compression via Group Merging and Layer Semantic Alignment}},
  author = {Jiwei Tang and Zhicheng Zhang and Shunlong Wu and Jingheng Ye and Lichen Bai and Zitai Wang and Tingwei Lu and Lin Hai and Yiming Zhao and Hai-Tao Zheng and Hong-Gee Kim},
  year = {2026},
  booktitle = {Proceedings of the 64th Annual Meeting of the Association for Computational Linguistics (Volume 1: Long Papers)},
  doi = {10.18653/v1/2026.acl-long.1324},
  pages = {28690--28704},
  url = {https://aclanthology.org/2026.acl-long.1324/}
}

@article{rw_sparse_transformer,
  title = {{Generating Long Sequences with Sparse Transformers}},
  author = {Child, Rewon and Gray, Scott and Radford, Alec and Sutskever, Ilya},
  year = {2019},
  journal = {arXiv preprint arXiv:1904.10509},
  url = {https://arxiv.org/abs/1904.10509}
}

@article{rw_longformer,
  title = {{Longformer: The Long-Document Transformer}},
  author = {Beltagy, Iz and Peters, Matthew E. and Cohan, Arman},
  year = {2020},
  journal = {arXiv preprint arXiv:2004.05150},
  url = {https://arxiv.org/abs/2004.05150}
}

@inproceedings{rw_bigbird,
  title = {{Big Bird: Transformers for Longer Sequences}},
  author = {Zaheer, Manzil and Guruganesh, Guru and Dubey, Kumar Avinava and Ainslie, Joshua and Alberti, Chris and Ontanon, Santiago and Pham, Philip and Ravula, Anirudh and Wang, Qifan and Yang, Li and Ahmed, Amr},
  year = {2020},
  booktitle = {Advances in Neural Information Processing Systems},
  volume = {33},
  pages = {17283--17297},
  url = {https://proceedings.neurips.cc/paper_files/paper/2020/hash/c8512d142a2d849725f31a9a7a361ab9-Abstract.html}
}

@article{rw_mistral,
  title = {{Mistral 7B}},
  author = {Jiang, Albert Q. and Sablayrolles, Alexandre and Mensch, Arthur and Bamford, Chris and Chaplot, Devendra Singh and Casas, Diego de las and Bressand, Florian and Lengyel, Gianna and Lample, Guillaume and Saulnier, Lucile and Lavaud, Lélio Renard and Lachaux, Marie-Anne and Stock, Pierre and Scao, Teven Le and Lavril, Thibaut and Wang, Thomas and Lacroix, Timothée and Sayed, William El},
  year = {2023},
  journal = {arXiv preprint arXiv:2310.06825},
  url = {https://arxiv.org/abs/2310.06825}
}

@inproceedings{rw_minference,
  title = {{MInference 1.0: Accelerating Pre-filling for Long-Context LLMs via Dynamic Sparse Attention}},
  author = {Jiang, Huiqiang and Li, Yucheng and Zhang, Chengruidong and Wu, Qianhui and Luo, Xufang and Ahn, Surin and Han, Zhenhua and Abdi, Amir H. and Li, Dongsheng and Lin, Chin-Yew and Yang, Yuqing and Qiu, Lili},
  year = {2024},
  booktitle = {Advances in Neural Information Processing Systems},
  volume = {37},
  doi = {10.52202/079017-1663},
  pages = {52481--52515},
  url = {https://proceedings.neurips.cc/paper_files/paper/2024/hash/5dfbe6f5671e82c76841ba687a8a9ecb-Abstract-Conference.html}
}

@inproceedings{rw_seerattention,
  title = {{SeerAttention: Self-distilled Attention Gating for Efficient Long-context Prefilling}},
  author = {Gao, Yizhao and Zeng, Zhichen and Du, DaYou and Cao, Shijie and Zhou, Peiyuan and Qi, Jiaxing and Lai, Junjie and So, Hayden and Cao, Ting and Yang, Fan and Yang, Mao},
  year = {2025},
  booktitle = {Advances in Neural Information Processing Systems},
  volume = {38},
  doi = {10.52202/085713-1869},
  pages = {55846--55869},
  url = {https://proceedings.neurips.cc/paper_files/paper/2025/hash/50e9dbc4ab68d94f15261ddc26c8ca2b-Abstract-Conference.html}
}

@inproceedings{rw_nsa,
  title = {{Native Sparse Attention: Hardware-Aligned and Natively Trainable Sparse Attention}},
  author = {Jingyang Yuan and Huazuo Gao and Damai Dai and Junyu Luo and Liang Zhao and Zhengyan Zhang and Zhenda Xie and Yuxing Wei and Lean Wang and Zhiping Xiao and Yuqing Wang and Chong Ruan and Ming Zhang and Wenfeng Liang and Wangding Zeng},
  year = {2025},
  booktitle = {Proceedings of the 63rd Annual Meeting of the Association for Computational Linguistics (Volume 1: Long Papers)},
  doi = {10.18653/v1/2025.acl-long.1126},
  pages = {23078--23097},
  url = {https://aclanthology.org/2025.acl-long.1126/}
}

@inproceedings{rw_spargeattn,
  title = {{SpargeAttention: Accurate and Training-free Sparse Attention Accelerating Any Model Inference}},
  author = {Jintao Zhang and Chendong Xiang and Haofeng Huang and Jia Wei and Haocheng Xi and Jun Zhu and Jianfei Chen},
  year = {2025},
  booktitle = {International Conference on Machine Learning},
  publisher = {PMLR},
  pages = {76397--76413},
  url = {https://proceedings.mlr.press/v267/zhang25ch.html}
}

@inproceedings{rw_moba,
  title = {{MoBA: Mixture of Block Attention for Long-Context LLMs}},
  author = {Lu, Enzhe and Jiang, Zhejun and Liu, Jingyuan and Du, Yulun and Jiang, Tao and Hong, Chao and Liu, Shaowei and He, Weiran and Yuan, Enming and Wang, Yuzhi and Huang, Zhiqi and Yuan, Huan and Xu, Suting and Xu, Xinran and Lai, Guokun and Chen, Yanru and Zheng, Huabin and Yan, Junjie and Su, Jianlin and Wu, Yuxin and Zhang, Yutao and Yang, Zhilin and Zhou, Xinyu and Zhang, Mingxing and Qiu, Jiezhong},
  year = {2025},
  booktitle = {Advances in Neural Information Processing Systems},
  volume = {38},
  doi = {10.52202/085713-0602},
  pages = {17790--17815},
  url = {https://proceedings.neurips.cc/paper_files/paper/2025/hash/19eae75beed66321d62272e794a9c2ac-Abstract-Conference.html}
}

@article{rw_deepseekv32,
  title = {{DeepSeek-V3.2: Pushing the Frontier of Open Large Language Models}},
  author = {{DeepSeek-AI} and Liu, Aixin and Mei, Aoxue and Lin, Bangcai and Xue, Bing and Wang, Bingxuan and Xu, Bingzheng and Wu, Bochao and Zhang, Bowei and others},
  year = {2025},
  journal = {arXiv preprint arXiv:2512.02556},
  url = {https://arxiv.org/abs/2512.02556}
}

@article{rw_qsa,
  title = {{On the Design of Qwen3.8-Next Architecture: Evaluation, Efficiency, and Training Stability}},
  author = {Qiu, Zihan and Wang, Zekun and Li, Xiao and Li, Yanpeng and Xu, Yang and Wang, Yixuan and Zhang, Huaqing and Men, Rui and Mao, Bochao and others},
  year = {2026},
  journal = {arXiv preprint arXiv:2608.30320},
  url = {https://arxiv.org/abs/2608.30320}
}

@article{rw_deepseekv4,
  title = {{DeepSeek-V4: Towards Highly Efficient Million-Token Context Intelligence}},
  author = {{DeepSeek-AI} and Xu, Anyi and Lin, Bangcai and Xue, Bing and Wang, Bingxuan and Xu, Bingzheng and Wu, Bochao and Zhang, Bowei and Lin, Chaofan and others},
  year = {2026},
  journal = {arXiv preprint arXiv:2606.19348},
  url = {https://arxiv.org/abs/2606.19348}
}

@article{rw_deepseekv41,
  title = {{DeepSeek-V4.1-Flash: Pushing the Limits of KV Cache Compression}},
  author = {{DeepSeek-AI}},
  year = {2026},
  journal = {arXiv preprint arXiv:2609.19969},
  url = {https://arxiv.org/abs/2609.19969}
}

@article{rw_hils,
  title = {{Hierarchical Sparse Attention Done Right: Toward Infinite Context Modeling}},
  author = {Hu, Xiang and Wei, Xinyu and Gu, Hao and Zhang, Minshen and Liang, Tian and Li, Huayang and Zhu, Lei and Wang, Yan and Han, Sirui and Bai, Yushi and Tu, Kewei and Mi, Haitao and Liang, Leo},
  year = {2026},
  journal = {arXiv preprint arXiv:2607.02980},
  url = {https://arxiv.org/abs/2607.02980}
}

@article{rw_hysparse,
  title = {{HySparse: A Hybrid Sparse Attention Architecture with Oracle Token Selection and KV Cache Sharing}},
  author = {Gao, Yizhao and Wei, Jianyu and Zhang, Qihao and Cheng, Yu and Chen, Shimao and Tang, Zhengju and Jiang, Zihan and Song, Yifan and Zhang, Hailin and Zhao, Liang and Yang, Bo and Wang, Gang and Cao, Shijie and Luo, Fuli},
  year = {2026},
  journal = {arXiv preprint arXiv:2602.03560},
  url = {https://arxiv.org/abs/2602.03560}
}

@article{rw_sas,
  title = {{SAS: Simple Attention Sparsification via End-to-End Optimization of Context Ranking}},
  author = {Li, Zhiwei and Zhu, Lei and Gu, Hao and Hu, Xiang and Wang, Yan and Mi, Haitao and Han, Sirui and Liang, Leo and Guo, Zhijiang},
  year = {2026},
  journal = {arXiv preprint arXiv:2609.13141},
  url = {https://arxiv.org/abs/2609.13141}
}

@article{rw_gemma3,
  title = {{Gemma 3 Technical Report}},
  author = {{Gemma Team} and Kamath, Aishwarya and Ferret, Johan and Pathak, Shreya and Vieillard, Nino and Merhej, Ramona and Perrin, Sarah and Matejovicova, Tatiana and Ramé, Alexandre and others},
  year = {2025},
  journal = {arXiv preprint arXiv:2503.19786},
  url = {https://arxiv.org/abs/2503.19786}
}

@article{rw_mimo2,
  title = {{MiMo-V2-Flash Technical Report}},
  author = {{Xiaomi LLM-Core Team} and Xiao, Bangjun and Xia, Bingquan and Yang, Bo and Gao, Bofei and Shen, Bowen and Zhang, Chen and He, Chenhong and Lou, Chiheng and others},
  year = {2026},
  journal = {arXiv preprint arXiv:2601.02780},
  url = {https://arxiv.org/abs/2601.02780}
}

@inproceedings{rw_xattention,
  title = {{XAttention: Block Sparse Attention with Antidiagonal Scoring}},
  author = {Ruyi Xu and Guangxuan Xiao and Haofeng Huang and Junxian Guo and Song Han},
  year = {2025},
  booktitle = {International Conference on Machine Learning},
  publisher = {PMLR},
  pages = {69819--69831},
  url = {https://proceedings.mlr.press/v267/xu25ag.html}
}

@article{rw_ssa,
  title = {{Simplified Sparse Attention via Gist Tokens}},
  author = {Mao, Yuzhen and Li, Michael Y. and Fox, Emily B.},
  year = {2026},
  journal = {arXiv preprint arXiv:2604.20920},
  url = {https://arxiv.org/abs/2604.20920v2}
}

@inproceedings{rw_linear_attention,
  title = {{Transformers are RNNs: Fast Autoregressive Transformers with Linear Attention}},
  author = {Angelos Katharopoulos and Apoorv Vyas and Nikolaos Pappas and François Fleuret},
  year = {2020},
  booktitle = {International Conference on Machine Learning},
  publisher = {PMLR},
  pages = {5156--5165},
  url = {https://proceedings.mlr.press/v119/katharopoulos20a.html}
}

@inproceedings{rw_h3,
  title = {{Hungry Hungry Hippos: Towards Language Modeling with State Space Models}},
  author = {Fu, Daniel Y. and Dao, Tri and Saab, Khaled K. and Thomas, Armin W. and Rudra, Atri and Ré, Christopher},
  year = {2023},
  booktitle = {International Conference on Learning Representations},
  url = {https://openreview.net/forum?id=COZDy0WYGg}
}

@article{rw_retnet,
  title = {{Retentive Network: A Successor to Transformer for Large Language Models}},
  author = {Sun, Yutao and Dong, Li and Huang, Shaohan and Ma, Shuming and Xia, Yuqing and Xue, Jilong and Wang, Jianyong and Wei, Furu},
  year = {2023},
  journal = {arXiv preprint arXiv:2307.08621},
  url = {https://arxiv.org/abs/2307.08621}
}

@inproceedings{rw_mamba,
  title = {{Mamba: Linear-Time Sequence Modeling with Selective State Spaces}},
  author = {Albert Gu and Tri Dao},
  year = {2024},
  booktitle = {Conference on Language Modeling},
  url = {https://openreview.net/forum?id=tEYskw1VY2}
}

@inproceedings{rw_gla,
  title = {{Gated Linear Attention Transformers with Hardware-Efficient Training}},
  author = {Songlin Yang and Bailin Wang and Yikang Shen and Rameswar Panda and Yoon Kim},
  year = {2024},
  booktitle = {International Conference on Machine Learning},
  publisher = {PMLR},
  pages = {56501--56523},
  url = {https://proceedings.mlr.press/v235/yang24ab.html}
}

@inproceedings{rw_jamba,
  title = {{Jamba: Hybrid Transformer-Mamba Language Models}},
  author = {Lenz, Barak and Lieber, Opher and Arazi, Alan and Bergman, Amir and Manevich, Avshalom and Peleg, Barak and Aviram, Ben and Almagor, Chen and Fridman, Clara and others},
  year = {2025},
  booktitle = {International Conference on Learning Representations},
  pages = {67959--67984},
  url = {https://proceedings.iclr.cc/paper_files/paper/2025/hash/a9ed43fa31dc8b4a7d7a673d713dcb5f-Abstract-Conference.html}
}

@inproceedings{rw_mamba2,
  title = {{Transformers are SSMs: Generalized Models and Efficient Algorithms Through Structured State Space Duality}},
  author = {Tri Dao and Albert Gu},
  year = {2024},
  booktitle = {International Conference on Machine Learning},
  publisher = {PMLR},
  pages = {10041--10071},
  url = {https://proceedings.mlr.press/v235/dao24a.html}
}

@article{rw_mamba_hybrid,
  title = {{An Empirical Study of Mamba-based Language Models}},
  author = {Waleffe, Roger and Byeon, Wonmin and Riach, Duncan and Norick, Brandon and Korthikanti, Vijay and Dao, Tri and Gu, Albert and Hatamizadeh, Ali and Singh, Sudhakar and Narayanan, Deepak and Kulshreshtha, Garvit and Singh, Vartika and Casper, Jared and Kautz, Jan and Shoeybi, Mohammad and Catanzaro, Bryan},
  year = {2024},
  journal = {arXiv preprint arXiv:2406.07887},
  url = {https://arxiv.org/abs/2406.07887}
}

@article{rw_zamba,
  title = {{Zamba: A Compact 7B SSM Hybrid Model}},
  author = {Glorioso, Paolo and Anthony, Quentin and Tokpanov, Yury and Whittington, James and Pilault, Jonathan and Ibrahim, Adam and Millidge, Beren},
  year = {2024},
  journal = {arXiv preprint arXiv:2405.16712},
  url = {https://arxiv.org/abs/2405.16712}
}

@inproceedings{rw_hymba,
  title = {{Hymba: A Hybrid-head Architecture for Small Language Models}},
  author = {Dong, Xin and Fu, Yonggan and Diao, Shizhe and Byeon, Wonmin and Chen, Zijia and Mahabaleshwarkar, Ameya and Liu, Shih-Yang and Van keirsbilck, Matthijs and Chen, Min-Hung and Suhara, Yoshi and Lin, Yingyan Celine and Kautz, Jan and Molchanov, Pavlo},
  year = {2025},
  booktitle = {International Conference on Learning Representations},
  pages = {97729--97755},
  url = {https://proceedings.iclr.cc/paper_files/paper/2025/hash/f32def07618040e540e0a6182e290562-Abstract-Conference.html}
}

@inproceedings{rw_gated_deltanet,
  title = {{Gated Delta Networks: Improving Mamba2 with Delta Rule}},
  author = {Yang, Songlin and Kautz, Jan and Hatamizadeh, Ali},
  year = {2025},
  booktitle = {International Conference on Learning Representations},
  pages = {29687--29707},
  url = {https://proceedings.iclr.cc/paper_files/paper/2025/hash/4904fad153f6434a7bcf04465d4be2cc-Abstract-Conference.html}
}

@article{rw_kimi_linear,
  title = {{Kimi Linear: An Expressive, Efficient Attention Architecture}},
  author = {{Kimi Team} and Zhang, Yu and Lin, Zongyu and Yao, Xingcheng and Hu, Jiaxi and Meng, Fanqing and Liu, Chengyin and Men, Xin and Yang, Songlin and others},
  year = {2025},
  journal = {arXiv preprint arXiv:2510.26692},
  url = {https://arxiv.org/abs/2510.26692}
}

@inproceedings{rw_samba,
  title = {{Samba: Simple Hybrid State Space Models for Efficient Unlimited Context Language Modeling}},
  author = {Ren, Liliang and Liu, Yang and Lu, Yadong and shen, yelong and Liang, Chen and Chen, Weizhu},
  year = {2025},
  booktitle = {International Conference on Learning Representations},
  pages = {53551--53575},
  url = {https://proceedings.iclr.cc/paper_files/paper/2025/hash/84a7fc24ed52e8eff514c33e8ac76ea3-Abstract-Conference.html}
}

@inproceedings{rw_h2o,
  title = {{H2O: Heavy-Hitter Oracle for Efficient Generative Inference of Large Language Models}},
  author = {Zhang, Zhenyu and Sheng, Ying and Zhou, Tianyi and Chen, Tianlong and Zheng, Lianmin and Cai, Ruisi and Song, Zhao and Tian, Yuandong and Ré, Christopher and Barrett, Clark and Wang, Zhangyang "Atlas" and Chen, Beidi},
  year = {2023},
  booktitle = {Advances in Neural Information Processing Systems},
  volume = {36},
  doi = {10.52202/075280-1506},
  pages = {34661--34710},
  url = {https://proceedings.neurips.cc/paper_files/paper/2023/hash/6ceefa7b15572587b78ecfcebb2827f8-Abstract-Conference.html}
}

@inproceedings{rw_streamingllm,
  title = {{Efficient Streaming Language Models with Attention Sinks}},
  author = {Xiao, Guangxuan and Tian, Yuandong and Chen, Beidi and Han, Song and Lewis, Mike},
  year = {2024},
  booktitle = {International Conference on Learning Representations},
  pages = {21875--21895},
  url = {https://proceedings.iclr.cc/paper_files/paper/2024/hash/5e5fd18f863cbe6d8ae392a93fd271c9-Abstract-Conference.html}
}

@inproceedings{rw_pagedattention,
  title = {{Efficient Memory Management for Large Language Model Serving with PagedAttention}},
  author = {Woosuk Kwon and Zhuohan Li and Siyuan Zhuang and Ying Sheng and Lianmin Zheng and Cody Hao Yu and Joseph E. Gonzalez and Hao Zhang and Ion Stoica},
  year = {2023},
  booktitle = {Proceedings of the 29th Symposium on Operating Systems Principles},
  doi = {10.1145/3600006.3613165},
  pages = {611--626},
  url = {https://dl.acm.org/doi/10.1145/3600006.3613165}
}

@inproceedings{rw_kivi,
  title = {{KIVI: A Tuning-Free Asymmetric 2bit Quantization for KV Cache}},
  author = {Zirui Liu and Jiayi Yuan and Hongye Jin and Shaochen Zhong and Zhaozhuo Xu and Vladimir Braverman and Beidi Chen and Xia Hu},
  year = {2024},
  booktitle = {International Conference on Machine Learning},
  publisher = {PMLR},
  pages = {32332--32344},
  url = {https://proceedings.mlr.press/v235/liu24bz.html}
}

@inproceedings{rw_snapkv,
  title = {{SnapKV: LLM Knows What You are Looking for Before Generation}},
  author = {Li, Yuhong and Huang, Yingbing and Yang, Bowen and Venkitesh, Bharat and Locatelli, Acyr and Ye, Hanchen and Cai, Tianle and Lewis, Patrick and Chen, Deming},
  year = {2024},
  booktitle = {Advances in Neural Information Processing Systems},
  volume = {37},
  doi = {10.52202/079017-0722},
  pages = {22947--22970},
  url = {https://proceedings.neurips.cc/paper_files/paper/2024/hash/28ab418242603e0f7323e54185d19bde-Abstract-Conference.html}
}

@inproceedings{rw_quest,
  title = {{QUEST: Query-Aware Sparsity for Efficient Long-Context LLM Inference}},
  author = {Jiaming Tang and Yilong Zhao and Kan Zhu and Guangxuan Xiao and Baris Kasikci and Song Han},
  year = {2024},
  booktitle = {International Conference on Machine Learning},
  publisher = {PMLR},
  pages = {47901--47911},
  url = {https://proceedings.mlr.press/v235/tang24l.html}
}

@inproceedings{hendrycksmeasuring,
  title = {{Measuring Massive Multitask Language Understanding}},
  author = {Hendrycks, Dan and Burns, Collin and Basart, Steven and Zou, Andy and Mazeika, Mantas and Song, Dawn and Steinhardt, Jacob},
  year = {2021},
  booktitle = {International Conference on Learning Representations},
  url = {https://openreview.net/forum?id=d7KBjmI3GmQ}
}

@inproceedings{li2024cmmlu,
  title = {{CMMLU: Measuring massive multitask language understanding in Chinese}},
  author = {Haonan Li and Yixuan Zhang and Fajri Koto and Yifei Yang and Hai Zhao and Yeyun Gong and Nan Duan and Timothy Baldwin},
  year = {2024},
  booktitle = {Findings of the Association for Computational Linguistics: ACL 2024},
  doi = {10.18653/v1/2024.findings-acl.671},
  pages = {11260--11285},
  url = {https://aclanthology.org/2024.findings-acl.671/}
}

@inproceedings{huang2023c,
  title = {{C-Eval: A Multi-Level Multi-Discipline Chinese Evaluation Suite for Foundation Models}},
  author = {Huang, Yuzhen and Bai, Yuzhuo and Zhu, Zhihao and Zhang, Junlei and Zhang, Jinghan and Su, Tangjun and Liu, Junteng and Lv, Chuancheng and Zhang, Yikai and lei, jiayi and Fu, Yao and Sun, Maosong and He, Junxian},
  year = {2023},
  booktitle = {Advances in Neural Information Processing Systems},
  volume = {36},
  doi = {10.52202/075280-2749},
  pages = {62991--63010},
  url = {https://proceedings.neurips.cc/paper_files/paper/2023/hash/c6ec1844bec96d6d32ae95ae694e23d8-Abstract-Datasets_and_Benchmarks.html}
}

@inproceedings{wang2024mmlu,
  title = {{MMLU-Pro: A More Robust and Challenging Multi-Task Language Understanding Benchmark}},
  author = {Wang, Yubo and Ma, Xueguang and Zhang, Ge and Ni, Yuansheng and Chandra, Abhranil and Guo, Shiguang and Ren, Weiming and Arulraj, Aaran and He, Xuan and Jiang, Ziyan and Li, Tianle and Ku, Max and Wang, Kai and Zhuang, Alex and Fan, Rongqi and Yue, Xiang and Chen, Wenhu},
  year = {2024},
  booktitle = {Advances in Neural Information Processing Systems},
  volume = {37},
  doi = {10.52202/079017-3018},
  pages = {95266--95290},
  url = {https://proceedings.neurips.cc/paper_files/paper/2024/hash/ad236edc564f3e3156e1b2feafb99a24-Abstract-Datasets_and_Benchmarks_Track.html}
}

@article{cobbe2021training,
  title = {{Training Verifiers to Solve Math Word Problems}},
  author = {Cobbe, Karl and Kosaraju, Vineet and Bavarian, Mohammad and Chen, Mark and Jun, Heewoo and Kaiser, Lukasz and Plappert, Matthias and Tworek, Jerry and Hilton, Jacob and Nakano, Reiichiro and Hesse, Christopher and Schulman, John},
  year = {2021},
  journal = {arXiv preprint arXiv:2110.14168},
  url = {https://arxiv.org/abs/2110.14168}
}

@article{wei2023cmath,
  title = {{CMATH: Can Your Language Model Pass Chinese Elementary School Math Test?}},
  author = {Wei, Tianwen and Luan, Jian and Liu, Wei and Dong, Shuang and Wang, Bin},
  year = {2023},
  journal = {arXiv preprint arXiv:2306.16636},
  url = {https://arxiv.org/abs/2306.16636}
}

@inproceedings{hendrycksmath2021,
  title = {{Measuring Mathematical Problem Solving With the MATH Dataset}},
  author = {Hendrycks, Dan and Burns, Collin and Kadavath, Saurav and Arora, Akul and Basart, Steven and Tang, Eric and Song, Dawn and Steinhardt, Jacob},
  year = {2021},
  booktitle = {Proceedings of the Neural Information Processing Systems Track on Datasets and Benchmarks},
  volume = {1},
  url = {https://datasets-benchmarks-proceedings.neurips.cc/paper_files/paper/2021/hash/be83ab3ecd0db773eb2dc1b0a17836a1-Abstract-round2.html}
}

@article{austin2021program,
  title = {{Program Synthesis with Large Language Models}},
  author = {Austin, Jacob and Odena, Augustus and Nye, Maxwell and Bosma, Maarten and Michalewski, Henryk and Dohan, David and Jiang, Ellen and Cai, Carrie and Terry, Michael and Le, Quoc and Sutton, Charles},
  year = {2021},
  journal = {arXiv preprint arXiv:2108.07732},
  url = {https://arxiv.org/abs/2108.07732}
}

@article{chen2021evaluating,
  title = {{Evaluating Large Language Models Trained on Code}},
  author = {Chen, Mark and Tworek, Jerry and Jun, Heewoo and Yuan, Qiming and Pinto, Henrique Ponde de Oliveira and Kaplan, Jared and Edwards, Harri and Burda, Yuri and Joseph, Nicholas and others},
  year = {2021},
  journal = {arXiv preprint arXiv:2107.03374},
  url = {https://arxiv.org/abs/2107.03374}
}

@misc{martin2023NIAH,
  title = {{Needle In A Haystack - Pressure Testing LLMs}},
  author = {Greg Kamradt},
  year = {2023},
  howpublished = {Official website / repository},
  note = {Evaluation repository, accessed September 17, 2026},
  url = {https://github.com/gkamradt/LLMTest_NeedleInAHaystack}
}

@inproceedings{peng2024eagle,
  title = {{Eagle and Finch: {RWKV} with Matrix-Valued States and Dynamic Recurrence}},
  author = {Bo Peng and Daniel Goldstein and Quentin Gregory Anthony and Alon Albalak and Eric Alcaide and Stella Biderman and Eugene Cheah and Teddy Ferdinan and Kranthi Kiran GV and Haowen Hou and Satyapriya Krishna and McClelland, Jr., Ronald and Niklas Muennighoff and Fares Obeid and Atsushi Saito and Guangyu Song and Haoqin Tu and Ruichong Zhang and Bingchen Zhao and Qihang Zhao and Jian Zhu and Rui-Jie Zhu},
  year = {2024},
  booktitle = {Conference on Language Modeling},
  url = {https://openreview.net/forum?id=soz1SEiPeq}
}

@article{botev2024recurrentgemma,
  title = {{RecurrentGemma: Moving Past Transformers for Efficient Open Language Models}},
  author = {Botev, Aleksandar and De, Soham and Smith, Samuel L and Fernando, Anushan and Muraru, George-Cristian and Haroun, Ruba and Berrada, Leonard and Pascanu, Razvan and Sessa, Pier Giuseppe and others},
  year = {2024},
  journal = {arXiv preprint arXiv:2404.07839},
  url = {https://arxiv.org/abs/2404.07839}
}

@inproceedings{chevalier2023adapting,
  title = {{Adapting Language Models to Compress Contexts}},
  author = {Alexis Chevalier and Alexander Wettig and Anirudh Ajith and Danqi Chen},
  year = {2023},
  booktitle = {Proceedings of the 2023 Conference on Empirical Methods in Natural Language Processing},
  doi = {10.18653/v1/2023.emnlp-main.232},
  pages = {3829--3846},
  url = {https://aclanthology.org/2023.emnlp-main.232/}
}

@inproceedings{li2025admtree,
  title = {{AdmTree: Compressing Lengthy Context with Adaptive Semantic Trees}},
  author = {Li, Yangning and Chen, Shaoshen and Li, Yinghui and Chen, Yankai and Zheng, Hai-Tao and Wang, Hui and Jiang, Wenhao and Yu, Philip S},
  year = {2025},
  booktitle = {Advances in Neural Information Processing Systems},
  volume = {38},
  doi = {10.52202/085713-1348},
  pages = {40389--40415},
  url = {https://proceedings.neurips.cc/paper_files/paper/2025/hash/39a397e1721555ddfa83cea2de225760-Abstract-Conference.html}
}

@inproceedings{rw_flashattention,
  title = {{FlashAttention: Fast and Memory-Efficient Exact Attention with IO-Awareness}},
  author = {Dao, Tri and Fu, Dan and Ermon, Stefano and Rudra, Atri and Ré, Christopher},
  year = {2022},
  booktitle = {Advances in Neural Information Processing Systems},
  volume = {35},
  doi = {10.52202/068431-1189},
  pages = {16344--16359},
  url = {https://proceedings.neurips.cc/paper_files/paper/2022/hash/67d57c32e20fd0a7a302cb81d36e40d5-Abstract-Conference.html}
}

\end{document}